# Modeling and Implementation of Quadcopter Autonomous Flight Based on Alternative Methods to Determine Propeller Parameters

Gene Patrick Rible*, Nicolette Ann Arriola, Manuel Ramos Jr.

*Electrical and Electronics Engineering Institute, College of Engineering, University of the Philippines, 1101, Philippines*

ABSTRACT

*To properly simulate and implement a quadcopter flight control for intended load and flight conditions, the quadcopter model must have parameters on various relationships including propeller thrust-torque, thrust-PWM, and thrust–angular speed to a certain level of accuracy. Thrust-torque modeling requires an expensive reaction torque measurement sensor. In the absence of sophisticated equipment, the study comes up with alternative methods to complete the quadcopter model. The study also presents a method of modeling the rotational aerodynamic drag on the quadcopter. Although the resulting model of the reaction torque generated by the quadcopter's propellers and the model of the drag torque acting on the quadcopter body that are derived using the methods in this study may not yield the true values of these quantities, the experimental modeling techniques presented in this work ensure that the derived dynamic model for the quadcopter will nevertheless behave identically with the true model for the quadcopter. The derived dynamic model is validated by basic flight controller simulation and actual flight implementation. The model is used as basis for a quadcopter design, which eventually is used for test purposes of basic flight control. This study serves as a baseline for fail-safe control of a quadcopter experiencing an unexpected motor failure.*

## 1. Introduction

This paper is an extension of work originally presented in the 6th International Conference on Control, Automation and Robotics [1].

Multicopters as unmanned aerial vehicles (UAVs) are gaining massive attention in research because of their broad military and civilian applications [2]. Compared with other flying vehicles, multicopters have unique capabilities such as static hovering, vertical take-off and landing (VTOL) [3], and the ability to switch between any two arbitrary directions very quickly any time [4]. The simplest, most efficient, and most economical multicopter is the quadcopter since it contains the minimum number of propellers required to fully control its attitude and position. The quadcopter's dynamic model is used as a basis in designing and adjusting controllers for best flight performance. This new work presents cheaper, alternative methods to model the quadcopter using only basic mechanical tools and electrical measuring devices. The simpler alternative modeling methods presented in this paper can be used to successfully operate the quadcopter in basic flight or even in dynamic fail-safe flight mode. In fact, the modeling methods in this paper were employed in another study [1] on fail-safe control if a motor failure were to occur. The complete methods presented in this work are able to capture the effect of air resistance and propeller torque on the quadcopter dynamics without requiring expensive equipment such as a reaction torque sensor. This work is thus very useful to small-scale researchers who have limited equipment but who nevertheless wish to perform research works on quadcopters.

## 2. Background Information

### 2.1. Quadcopter Kinematics

In geometry, a coordinate is a set of values that locates a point. A coordinate frame is a system that uses a minimum of $n$ coordinates to uniquely determine the position of points of geometric objects in an $n$-dimensional manifold such as the 3-dimensional Euclidean space. This is done by assigning an $n$-tuple of values to every point on the manifold. One coordinate frame is linearly transformed into another through rotations and translations [5]. Thus, every rotation and translation of a coordinate frame is a transformation into another coordinate frame and because the operations involved in this transformation are linear, this transformation may be completely captured by matrix multiplication for the rotation and matrix addition for the translation.

*Corresponding Author: Gene Patrick Rible, 0889 Lim Extension, Digos, PH +63 956 257 2997, gsrible@up.edu.ph

In this paper, characters in **boldface** are vector or matrix quantities. Consider now the quadcopter kinematic diagram in Figure 1.a. There are three rotation angles $\phi$, $\theta$, $\psi$ and three axial coordinates $\boldsymbol{x}$, $\boldsymbol{y}$, $\boldsymbol{z}$. Four propellers are arranged symmetrically about the quadcopter's center of mass (COM). Propellers 1 and 3 rotate opposite to the rotation of propellers 2 and 4. Three coordinate axes $\boldsymbol{x}, \boldsymbol{y}, \boldsymbol{z}$ are drawn along the quadcopter arms and with origin at the COM. The direction of roll, pitch, and yaw angles and their corresponding body frame angular velocity components are indicated by rotation arrows along the coordinate axes.

Since the motion of the Earth is relatively very slow compared with the motion of the quadcopter in flight, a non-rotating Earth may be assumed. Moreover, since the radius of curvature of the Earth at any point on its surface is very large in comparison with the distances traveled by a quadcopter in flight, a flat Earth may be further assumed.

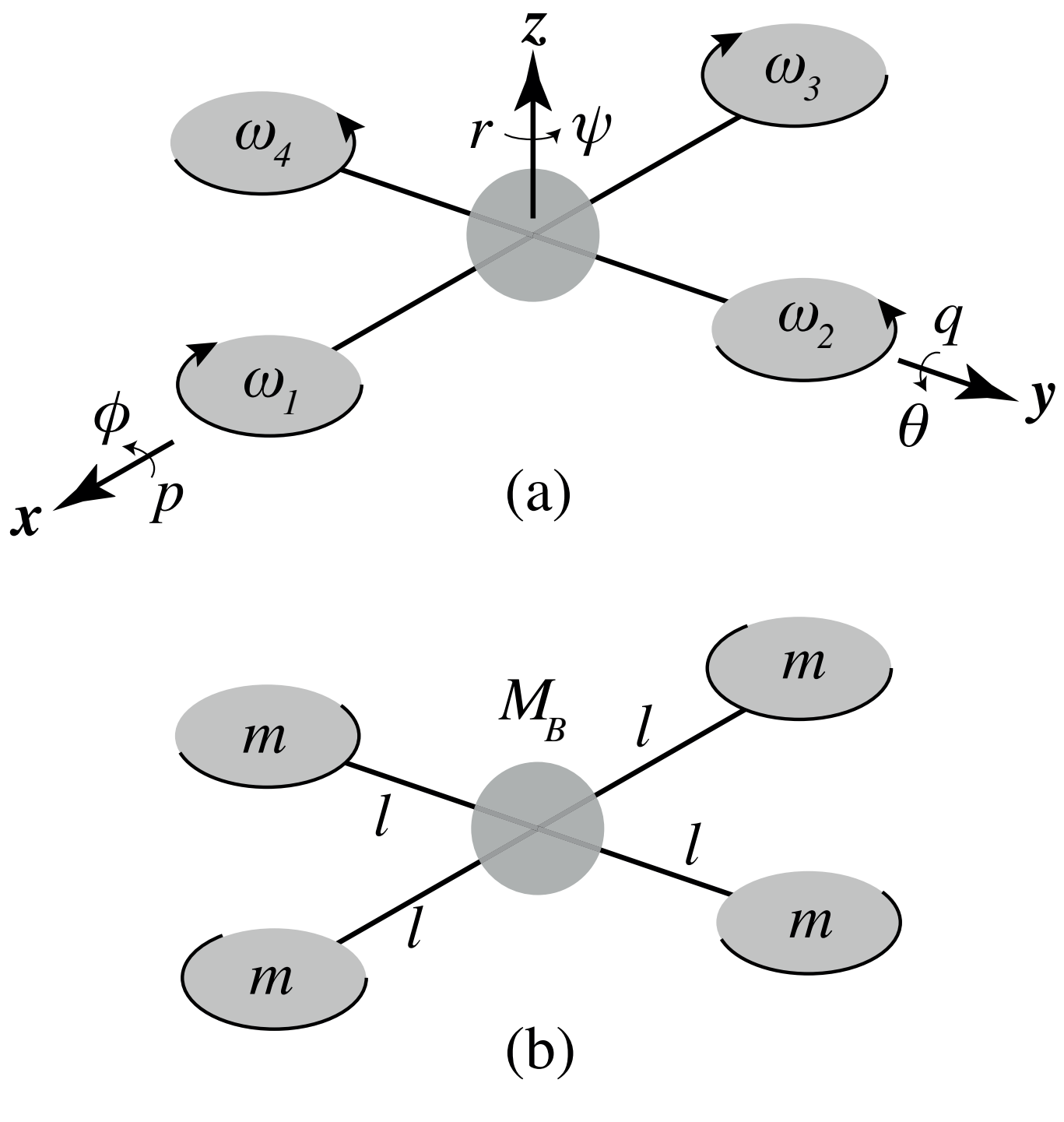

Figure 1: Quadcopter's (a) Kinematic [1] and (b) Dynamic Model

In this paper, a superscript indicates the coordinate frame with which an entity is defined. A rotation matrix that rotates $C^a$ to $C^b$ will be denoted by $\boldsymbol{R}_a^b$. Note that for a rotation matrix [5],

$$\boldsymbol{R}_b^a = {\boldsymbol{R}_a^b}^T = {\boldsymbol{R}_a^b}^{-1} \tag{1}$$

The quadcopter state variables are listed in Table 1. To control a quadcopter, state space readings are required. Under basic flight conditions, a state space that completely specifies the six degrees of freedom of a quadcopter and may be used for dynamic computations is $(\phi, \theta, \psi, p, q, r, x, y, z, \dot{x}, \dot{y}, \dot{z})$.

Table 1: Quadcopter State Variables [1]

| | |
|---|---|
| $x$ | inertial position along $\hat{\boldsymbol{x}}^i$ in $C^i$ |
| $y$ | inertial position along $\hat{\boldsymbol{y}}^i$ in $C^i$ |
| $z$ | altitude along $\hat{\boldsymbol{z}}^i$ in $C^i$ |
| $u$ | body frame velocity along $\hat{\boldsymbol{x}}^b$ in $C^b$ |
| $v$ | body frame velocity along $\hat{\boldsymbol{y}}^b$ in $C^b$ |
| $w$ | body frame velocity along $\hat{\boldsymbol{z}}^b$ in $C^b$ |
| $\phi$ | roll angle in $C^{v2}$ |
| $\theta$ | pitch angle in $C^{v1}$ |
| $\psi$ | yaw angle in $C^v$ |
| $p$ | roll angular velocity about $\hat{\boldsymbol{x}}^b$ in $C^b$ |
| $q$ | pitch angular velocity about $\hat{\boldsymbol{y}}^b$ in $C^b$ |
| $r$ | yaw angular velocity about $\hat{\boldsymbol{z}}^b$ in $C^b$ |

A generally accepted quadcopter dynamic model is shown in Figure 1.b. The four motors and propellers are assigned a mass $m$ centered at a distance $l$ from the quadcopter's geometric center such that all masses $m$ are 90° apart from each other with respect to the quadcopter's geometric center. The remaining body mass of the quadcopter excluding the four motors and propellers is $M_B$ which is centered at the quadcopter's geometric center. The centers of the four masses $m$ and of mass $M_B$ are assumed to be coplanar. The total mass of the quadcopter is thus

$$M = M_B + 4m \tag{2}$$

Since the quadcopter's dimensions are negligibly small in comparison with the Earth's radius of curvature, it is safe to assume that the center of gravity of any part of the quadcopter coincides with the object's COM.

### *2.2. Moment of Inertia*

Using the orthonormal $\boldsymbol{x}, \boldsymbol{y}, \boldsymbol{z}$ axes in Figure 1.a, a symmetric inertia matrix may be defined as follows:

$$\boldsymbol{J} \triangleq \begin{bmatrix} J_{xx} & -J_{xy} & -J_{xz} \\ -J_{yx} & J_{yy} & -J_{yz} \\ -J_{zx} & -J_{zy} & J_{zz} \end{bmatrix}$$

$$\triangleq \begin{bmatrix} \int (y^2+z^2)dm & -\int (xy)dm & -\int (xz)dm \\ -\int (yx)dm & \int (x^2+z^2)dm & -\int (yz)dm \\ -\int (zx)dm & -\int (zy)dm & \int (x^2+y^2)dm \end{bmatrix} \tag{3}$$

The diagonal components correspond to the axial moment of inertias (MOIs) of the quadcopter. The non-diagonal elements represent any inertial coupling that may be present between two axes of rotation. For a quadcopter that is symmetric about all three axes, the non-diagonal components of the inertia matrix are zero:

$$\boldsymbol{J}=\begin{bmatrix} J_{xx} & 0 & 0 \\ 0 & J_{yy} & 0 \\ 0 & 0 & J_{zz} \end{bmatrix} \quad (4)$$

*2.3. Drag Torque*

Due to air resistance, the drag vector $\boldsymbol{\tau}_d \triangleq \boldsymbol{\tau}_d^{\,b}=(0, 0, \tau_d)$ acting about $\hat{z}^b$ may further be added into the model especially since the quadcopter will exhibit spinning about this axis after losing one or more propellers as implemented in this project. This drag vector is experimentally modeled in [6] as linearly increasing with the yaw angular velocity:

$$\tau_d=\gamma r \quad (5)$$

where $\gamma$ is the drag coefficient. However, this is only true when the angular speed is very low and there is no turbulence [7]. In this work, it was experimentally found out that the drag torque is better approximated by

$$\tau_d=(\gamma_1 r^2+\gamma_2)\operatorname{sgn}(r) \quad (6)$$

where "sgn" denotes the signum function. The approximately quadratic relationship provided by $\gamma_1$, especially for high angular speeds, is theoretically supported [8]. Meanwhile, if the quadcopter is tied to a bearing, $\gamma_2$ is an offset provided by the rotational kinetic friction after the bearing's rotational static friction has been overcome. Without the bearing, $\gamma_2$ is simply set to zero.

*2.4. Rotational Dynamics*

The complete rotational dynamic model for the quadcopter is [5]

$$\dot{\boldsymbol{\omega}}_{b/i}=\boldsymbol{J}^{-1}\left(\boldsymbol{\tau}-\boldsymbol{\omega}_{b/i}\times\left\{\boldsymbol{J}\boldsymbol{\omega}_{b/i}+\boldsymbol{J}^P\boldsymbol{\Omega}\right\}-\boldsymbol{\tau}_d\right) \quad (7)$$

which in simplified matrix form is [1]

$$\begin{bmatrix}\dot{p}\\ \dot{q}\\ \dot{r}\end{bmatrix}=\begin{bmatrix}\frac{1}{J_{xx}}\tau_\phi\\ \frac{1}{J_{yy}}\tau_\theta\\ \frac{1}{J_{zz}}\tau_\psi\end{bmatrix}-\begin{bmatrix}\frac{J_{zz}-J_{yy}}{J_{xx}}qr\\ \frac{J_{xx}-J_{zz}}{J_{yy}}pr\\ \frac{J_{yy}-J_{xx}}{J_{zz}}pq\end{bmatrix}-\begin{bmatrix}\frac{J_{zz}^P}{J_{xx}}q\Omega\\ \frac{J_{zz}^P}{J_{yy}}p\Omega\\ 0\end{bmatrix}-\begin{bmatrix}0\\ 0\\ \frac{1}{J_{zz}}\tau_d\end{bmatrix} \quad (8)$$

where the algebraic sum of the propeller angular speeds is denoted by [9, 10]

$$\Omega=\sum_{i=1}^{4}(-1)^i\omega_i \quad (9)$$

## 3. Experimental Methods

*3.1. Hardware Design*

The quadcopter's skeletal body is the DJI Flame Wheel 450 with the commercial DJI E600 propulsion system [11]. The electronic speed controllers (ESCs) of this propulsion system work by current control and, as with most ESCs in the industry [12], vary the speed of the propellers depending on the on-time of the sent commands and not on the duty cycle. Thus, twice the duty cycle at twice the frequency would result in the same effective on-time and thus, the same propeller speed. This sets a limit on the maximum allowable frequency for the ESCs as the period of the signal sent must be greater than the minimum on-time required to turn on the propellers. The DJI E600 ESCs can receive motor signals between 30 Hz and 450 Hz. Note that this limit on the actuation frequency also sets a limit on the effectivity of the flight controller since even if the controller loop frequency is increased, a constant delay in the actuation will persist. In this project, the target loop frequency of the controller was set at 450 Hz.

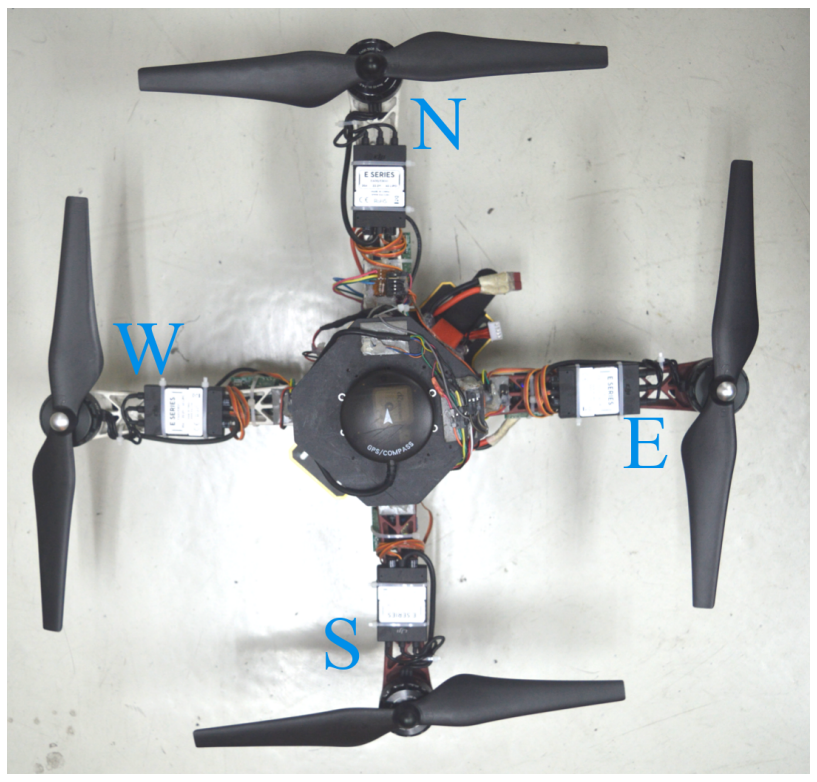

Figure 2: Quadcopter Platform Designed

A top view of the quadcopter for this project is shown in Figure 2. The letters N, E, W, and S indicate the designated North, East, West, and South directions respectively along the quadcopter's body frame axes. Since cross configuration is used in this paper, $\boldsymbol{x}^b$ is assigned to point along the North; $\boldsymbol{z}^b$ is assigned to point towards the reader; and $\boldsymbol{y}^b$ completes a right-handed coordinate frame. Note that the quadcopter is not symmetric about the $\boldsymbol{x}^b\boldsymbol{z}^b$ plane nor is it symmetric about the $\boldsymbol{y}^b\boldsymbol{z}^b$ plane. However, the quadcopter is symmetric about the $\boldsymbol{x}_1^{\,b}\boldsymbol{z}_b$ and $\boldsymbol{y}_1^{\,b}\boldsymbol{z}^b$ where $\boldsymbol{x}_1^{\,b}$ is $\boldsymbol{x}^b$ rotated 45° about $\boldsymbol{z}^b$ and $\boldsymbol{y}_1^{\,b}$ is $\boldsymbol{y}^b$ rotated 45° about $\boldsymbol{z}_b$. An axis formed by any plane, the $\boldsymbol{x}^b\boldsymbol{y}^b$ plane for example, and any plane orthogonal to $\boldsymbol{x}^b\boldsymbol{y}^b$ that bisects the body symmetrically is an *axis of symmetry* while such a bisecting plane is a *plane of symmetry*. If more than one axis of symmetry exists, then the intersection of any two axes of symmetry will be the body's *origin*. Assuming that the body's mass is evenly distributed and placing $\boldsymbol{x}^b\boldsymbol{y}^b$ such that $\boldsymbol{x}^b\boldsymbol{y}^b$ passes through the body's COM, the existence of one plane of symmetry means that rotation of the body about an axis created by any other plane orthogonal to $\boldsymbol{x}^b\boldsymbol{y}^b$ and passing through the origin will be as if the body is rotating about an axis of symmetry. This means that for this quadcopter, even though $\boldsymbol{x}^b$ and $\boldsymbol{y}^b$ are not axes of symmetry, rotating about $\boldsymbol{x}^b$ and $\boldsymbol{y}^b$ will be as if they are axes of symmetry and so Equation (4) will still hold using $C^b$.

*3.2. Quadcopter Modeling*

Methods to obtain the propeller parameters and subsequently complete the quadcopter's dynamic model are discussed in this section.

*3.2.1. Propeller Parameters*

Typically, a force sensor and a reaction torque sensor are necessary to come up with the propeller thrust-torque and thrust–pulse width modulation (PWM) relationships [13, 14]. However, these sophisticated sensors are highly expensive and not readily available. It is also extremely difficult, if not impossible, to find any force sensor or torque sensor product that will readily couple with a commercial UAV propeller to establish a working setup with enough airflow space for propeller modeling purposes. With these considerations in mind, an alternative setup was designed in this work to obtain such relationships for the model. The setup is shown in Figure 3.

The microcontroller sends PWM command to the propeller. As the propeller spins, a weighing scale measures the thrust while a tachometer measures the propeller angular speed. The reaction torque is computed from the voltage and current used to power the spinning propeller.

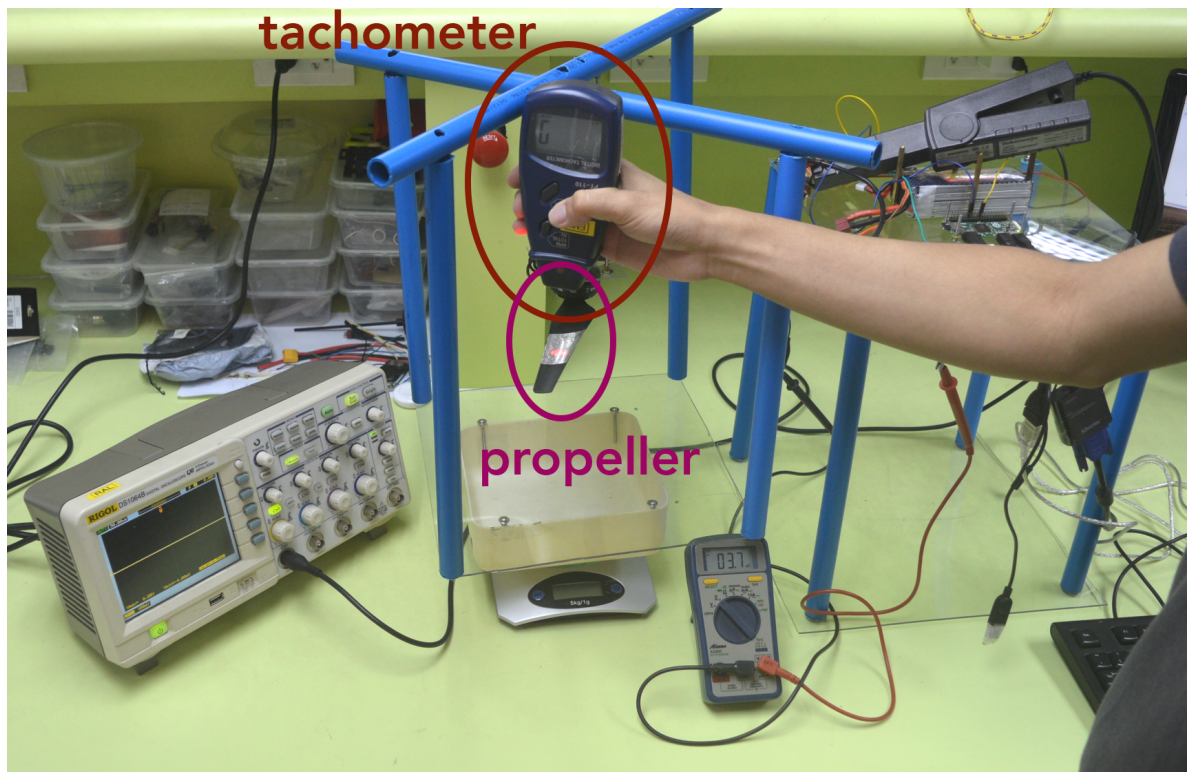

Figure 3: Alternative Propeller Modeling Method

The succeeding discussion on reaction torque appears in [1]. For every rotating propeller $i$, the produced thrust magnitude $f_i$, reaction torque magnitude $\tau_i$, and angular speed $\omega_i$ are related by [15]

$$f_i=k_f\omega_i^2 \tag{10}$$

$$\tau_i=k_\tau\omega_i^2 \tag{11}$$

$$f_i=k\tau_i \tag{12}$$

where $k_f$, $k_\tau$, $k$ are positive constants. At steady state, electrical torque $\tau_e$ equals mechanical torque $\tau_m$ [16]. The propeller exerts a torque in a direction opposite its sense of rotation [17]—a part of this is the reaction torque $\tau_\psi$ exerted by the spinning propeller onto the body frame while the remaining is dissipated by aerodynamic drag $\tau_{d_{prop}}$. Thus, at steady state,

$$\tau_e=\tau_m=\tau_\psi+\tau_{d_{prop}} \tag{13}$$

From [15],

$$\tau_\psi=k_\tau\omega^2 \tag{14}$$

where $k_\tau$ is a positive scalar. It was also shown in [18] that

$$\tau_{d_{prop}}=k_d\omega^2 \tag{15}$$

where $k_d$ is also a positive scalar. Since $\tau_\psi$ and $\tau_{d_{prop}}$ behave similarly with respect to $\omega$, it is impossible to distinguish between the two. We can set $\tau_{d_{prop}}=0$ so that

$$\tau_\psi=\tau_e \tag{16}$$

$\tau_{d_{prop}}$ is lumped together with the drag torque $\tau_d$ on the body frame via experimental modeling. This effectively eliminates whatever excess torque allotment that went into $\tau_\psi$.

#### 3.2.2. Axial Moments of Inertia

The axial MOIs of the quadcopter were obtained by suspending the quadcopter on three different axes as shown in Figure 4. The MOI of the quadcopter about the axis located at the pivot point is

$$J_P=Mgr\left(\frac{T}{2\pi}\right)^2 \tag{17}$$

where $T$ is the measured period of oscillation and $r$ is the distance between the COM and the pivot point; Equation (17) is derived in [19]. Note that $r$ is equal to the quadcopter arm length plus possibly a small additional length from the rope that tied the quadcopter to the pivot point. Using parallel axis theorem, the MOI of the quadcopter about the body frame axis passing through the quadcopter's COM is

$$J_C=J_P-Mr^2 \tag{18}$$

To accurately measure $T$, the quadcopter is filmed while swinging freely at 240 fps. The measured time intervals between 16 complete swings since the start of filming are averaged to yield an estimate of $T$. This is repeated three times. The average of the three trials is the assumed value for $T$. This averaging is important to protect the data from measurement errors. With this value, the quadcopter axial MOIs are then obtained using Equation (17) and Equation (18).

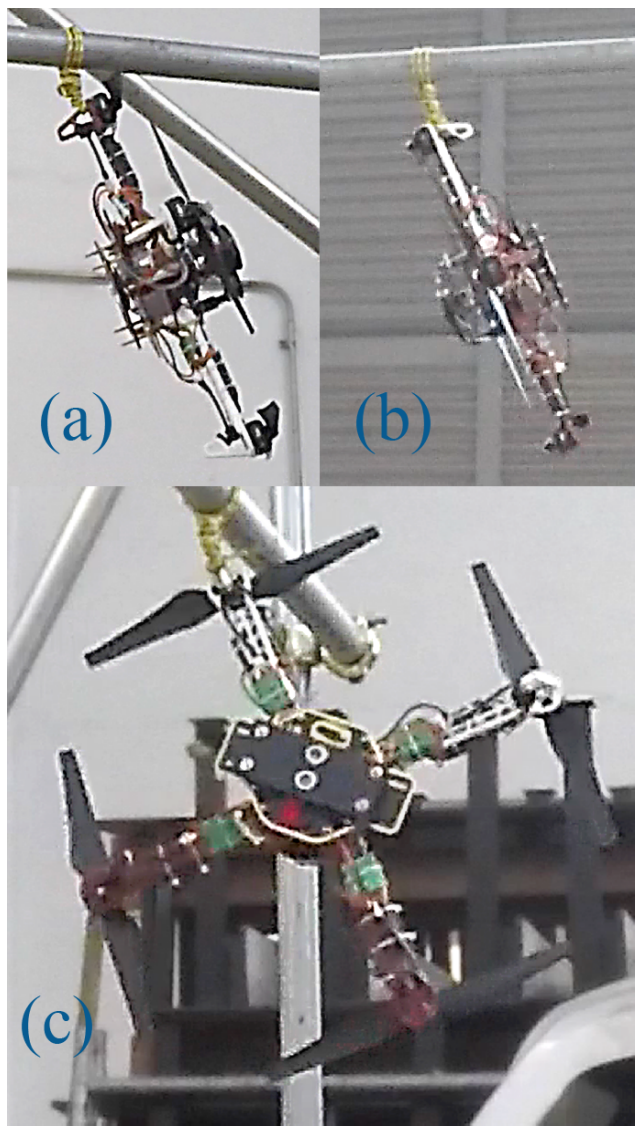

Figure 4: Measuring the Quadcopter MOI About (a) $\hat{x}^b$, (b) $\hat{y}^b$, and (c) $\hat{z}^b$

Meanwhile, to obtain the MOI of the motors and propeller blades about their axis of rotation, they were first weighed. Assuming that the propeller blade is a circular disk with an evenly distributed mass $m_b$ and radius $r_b$, its axial MOI can be shown to equal

$$J_{z,b}=\frac{1}{2}m_b {r_b}^2 \tag{19}$$

Assuming that the motor is a solid cylinder with an evenly distributed mass $m_m$ and radius $r_m$, its axial MOI can be shown to equal

$$J_{z,m}=\frac{1}{2}m_m {r_m}^2 \tag{20}$$

Since the propeller blade and motor have the same axis of rotation, their MOIs add up so that the propeller MOI about its axis of rotation is simply

$$J_{zz}^P=J_{z,b}+J_{z,m} \tag{21}$$

These were the same methods used in [6] to obtain $J_{zz}^P$. Since there are four propellers, four $J_{zz}^P$ were computed. Since the propellers are manufactured to be identical, their $J_{zz}^P$ values were averaged and the average value was taken as the $J_{zz}^P$ of any of the propellers. Again, this averaging was done since the motors and propellers are manufactured to be identical; this averaging would protect from experimental errors and remove biases.

### *3.2.3. Drag Coefficients*

The behavior of the drag torque due to air resistance may be experimentally determined by allowing the quadcopter body to rotate about $\hat{z}^b$ and applying different torques along the clockwise or counterclockwise direction. Increasing the applied torque will accelerate the rotation of the quadcopter about $\hat{z}^b$ until the drag torque due to the quadcopter's angular speed equals the applied torque, in which case the quadcopter's rotation about $\hat{z}^b$ reaches its steady state angular speed. Thus, different applied torques will correspond to different steady state angular speeds about $\hat{z}^b$ and the relationship of the applied torques (which become equal to the drag torques at steady state) with respect to the measured steady state angular speeds will determine the behavior of the drag torque due to air resistance. This behavior will be described by a quadratic equation consisting of drag coefficient(s). The same curve-fitting method based on experimental data plot, albeit using a simpler linear assumed model equation, was employed in [6].

Since air's viscosity increases with greater pressure and lower temperature, the air resistance will decrease with higher temperature and higher elevation (which lowers air pressure). It is possible to model the variation of the air resistance with respect to temperature and pressure if both the pressure and temperature of the environment can be controlled and maintained. However, without a closed room facility and equipment that maintains temperature and pressure, it is impossible to keep the temperature and pressure constant throughout the modeling of the air resistance versus angular velocity relationship. Moreover, a changing set of drag coefficients would imply a changing plant—this means that the periodic equilibrium solutions and controller decisions should also be dependent on temperature and pressure. Due to the lack of the aforementioned facility and equipment, and to simplify the problem, the air resistance versus the angular speed relationships were assumed constant and invariant to changes in pressure and temperature; this was also assumed in [6]. In a similar way, temperature and pressure could also affect the PWM to thrust relationship of the propellers. However, due again to the lack of the required facility and equipment to control ambient temperature and pressure and to simplify the problem, the PWM to thrust relationship was assumed constant and invariant to changes in pressure and temperature; this was also assumed in [13].

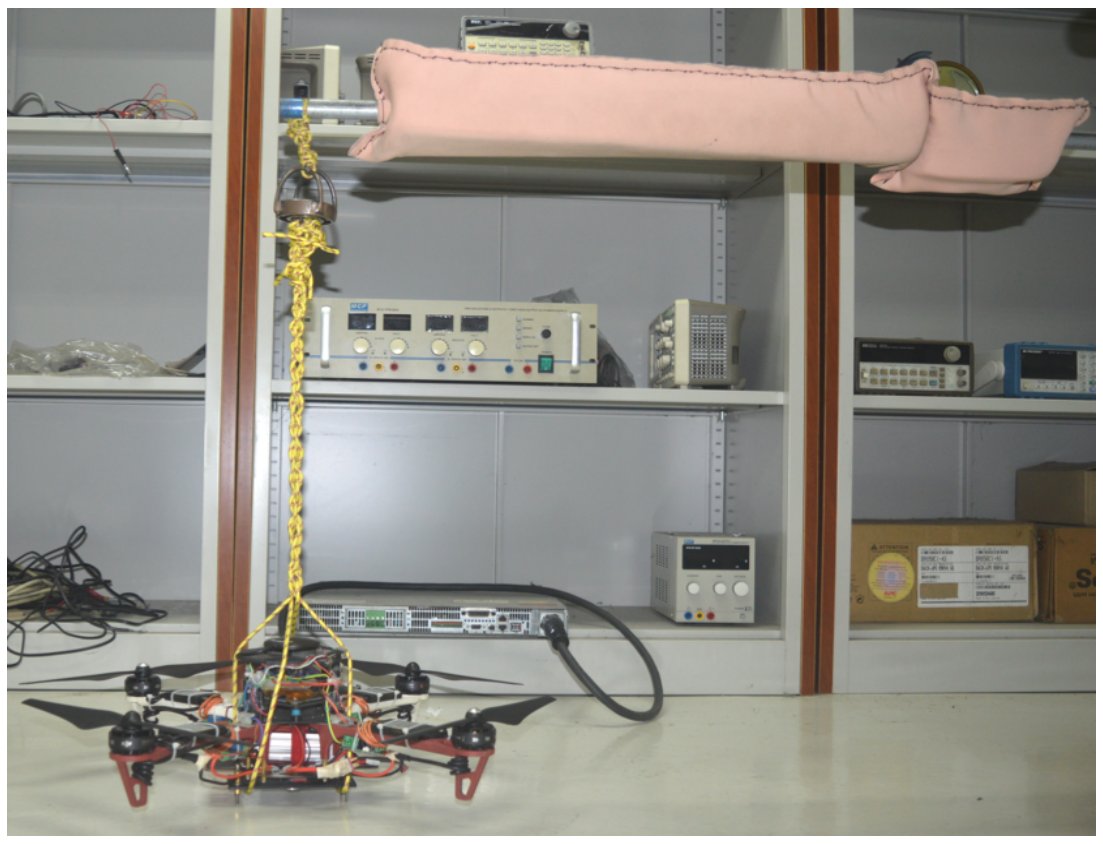

Figure 5: Rope and Bearing Setup for Air Resistance Characterization [1]

The drag torque on the quadcopter's body due to the air resistance was modeled by suspending the quadcopter along a bearing with a single axis of rotation as shown Figure 5. For the air resistance characterization setup, U-shaped metal rods were welded onto a circular bearing, resulting in an axially rotating pivot point. The quadcopter is made to rotate about $\hat{z}^b$ by commanding propellers 1 and 3 to exert a certain torque of the same value while disabling propellers 2 and 4. Each torque value will correspond to a steady state yaw angular velocity: the higher the torque value, the greater the magnitude of the steady state angular velocity of the quadcopter will be. The parametrization of the plot of the combined torque versus the steady state angular velocities will give the coefficients for the drag torque.

Because of metal welding, the metal junction areas expanded, making it hard for the bearing to rotate; this problem was solved using a rubbing compound to scrape off the expanded junction areas. A spherical ball bearing would allow for an omnidirectional rotation and would appear to be more preferable but under the same setup in Figure 5, the rotational friction it yielded was too high that it failed to rotate because of the weight of the quadcopter. In the end, the circular bearing was retained. Ideally, the rotational friction of the bearing would be unaffected by the mass that it carries. However, in reality, the heavier the mass, the greater the rotational kinetic and static friction. Dry lubricant was applied onto the bearing's interior to reduce the rotational friction.

## *3.3. Simulation of Basic Flight Controller with the Model Derived*

The proportional–integral–derivative (PID) controller output is

$$u(t)=k_p\left(y_d(t)-y(t)\right)+k_d\frac{d}{dt}\left(y_d(t)-y(t)\right)$$
$$+k_i\int\left(y_d(t)-y(t)\right)dt \quad (22)$$

where $u(t)$ is the control output; $y(t)$ is the state variable to be controlled; $y_d(t)$ is the desired reference value; $k_p$, $k_d$, and $k_i$ are non-negative constants to be tuned before running the controller. Implementing Equation (22) in a microcontroller or computer will result in a controller output that updates in discrete instances according to the controller loop frequency wherein the integral term becomes a running sum. Examining Equation (22), a sharp impulse called *derivative kick* can saturate the actuator and push away the system from the linear zone whenever a task requires $y_d(t)$ to change from one constant value to another [20]. Because of this, most of PID architectures use the derivative of $y(t)$ only. Thus,

$$u(t)=k_p\left(y_d(t)-y(t)\right)-k_d\frac{d}{dt}\left(y(t)\right)+k_i\int\left(y_d(t)-y(t)\right)dt \quad (23)$$

$$u(t)=k_p\left(y_d(t_k)-y(t_k)\right)-k_d\frac{y(t_k)-y(t_{k-1})}{t_k-t_{k-1}}$$
$$+k_i\sum_{j=1}^{k}\left(\left[y_d(t_j)-y(t_j)\right]\left[t_j-t_{j-1}\right]\right) \quad (24)$$

where Equation (24) is the discretized control law version of Equation (23) that can be implemented in microcontrollers.

The simulations were done at a controller frequency of 450 Hz; the higher the controller frequency, the closer Equation (24) will be to Equation (23). This frequency is chosen since it is the maximum signal frequency that can be sent to the motor ESCs.

The differential reference values used in tuning the basic flight controller in the simulations were ±2 m for the altitude, ±5° for the roll and pitch, and ±30° for the yaw. Furthermore, the settling period was set to 3 s; the desired number of oscillations for the altitude was set to 1; the desired number of oscillations for the yaw was set to 2; the desired number of oscillations for the roll and pitch was set to 11. The greater the desired number of oscillations for a variable, the stronger its proportional controller, the higher the variable overshoots, and the faster the system response will be for the variable.

In tuning the PID gains, $k_i$ is initially set to zero. Afterwards, $k_p$ is manually increased until the desired number of oscillations is reached. Then $k_d$ is manually increased until no oscillation remains. The allowable error was set to 1 cm for the altitude and 0.1° for the angles. Note that in this PID tuning method, a single initial overshoot in $y(t)$ is allowed. $k_i$ may be introduced in the end if steady state errors are to be removed. This tuning procedure was employed for the PID controllers of $\phi$, $\theta$, $\psi$, and $z$ with controller outputs $u_\phi$, $u_\theta$, $u_\psi$, and $u_z$. For the quadcopter,

$$\begin{bmatrix}f_1\\f_2\\f_3\\f_4\end{bmatrix}=\begin{bmatrix}B\\B\\B\\B\end{bmatrix}+\begin{bmatrix}u_z\\u_z\\u_z\\u_z\end{bmatrix}+\begin{bmatrix}0\\u_\phi\\0\\-u_\phi\end{bmatrix}+\begin{bmatrix}-u_\theta\\0\\u_\theta\\0\end{bmatrix}+\begin{bmatrix}u_\psi\\-u_\psi\\u_\psi\\-u_\psi\end{bmatrix} \quad (25)$$

where $B$ is the base weight which is set to $\frac{1}{4}Mg$ or one-fourth the total weight of the quadcopter; this is to ensure equal sharing of load among the four motors at hover. $u_z$, before being used in Equation (25), is first rescaled by dividing it by $\hat{z}^b\cdot\hat{z}^i=\cos(\phi)\cos(\theta)$ or the directional cosine between the quadcopter's body frame z-axis and the direction pointing opposite to gravity in order to take into account the quadcopter tilt which reduces the lift force [1].

## 4. Results and Discussion

### 4.1. Quadcopter Model and Simulation

The obtained parameters of the quadcopter used in this project are listed in Table 2. The standard acceleration due to gravity value is assumed for $g$. The rest of the parameters were obtained using methods in Section 3.

Table 2: Parameters of the Quadcopter

| Parameter | Value | Unit |
|---|---|---|
| $M$ | 1.645 | kg |
| $l$ | 0.2475 | m |
| $g$ | 9.80665 | m/s$^2$ |
| $J_{xx}$ | 0.014002764 | kg m$^2$ |
| $J_{yy}$ | 0.014267729 | kg m$^2$ |
| $J_{zz}$ | 0.029487252 | kg m$^2$ |
| $J_{zz}^P$ | 2.66838E–04 | kg m$^2$ |
| $\gamma_1$ | 4.86291E–04 | (N m)/(rad/s)$^2$ |
| $\gamma_2$ | 1.22958E–03 | N m |

For the propeller characteristics, the following models were experimentally obtained:

$$f_i=h_1(P_i-h_2)^2 \quad (26)$$

$$f_i=c_1\omega_i^2 \quad (27)$$

$$\tau_i=(g_1f_i+g_2)\,\mathrm{sgn}(f_i) \quad (28)$$

where $f_i$ is the force value in N, $P_i$ is a command value from 0 to 255 with 0 corresponding to 0% duty cycle and 255 corresponding to 100% duty cycle, $\tau_i$ is the torque value in N m, and $\omega_i$ is the propeller speed in $^{\mathrm{rad}}/_{\mathrm{s}}$ of propeller $i$. Note that Equation (27) and Equation (28) are based on Equation (10) and Equation (12).

Following the alternative method designed, as described in Section 3, the propeller thrust to PWM, propeller thrust to angular speed, and propeller thrust to torque relationships were obtained. This final set of parameters was also used in computing the total torque exerted by the propellers for the drag coefficients. Figure 6, Figure 7, and Figure 8 show plots of these combinations. The plots in yellow correspond to the final model whose parameters are the average of those of the obtained models for the four propellers.

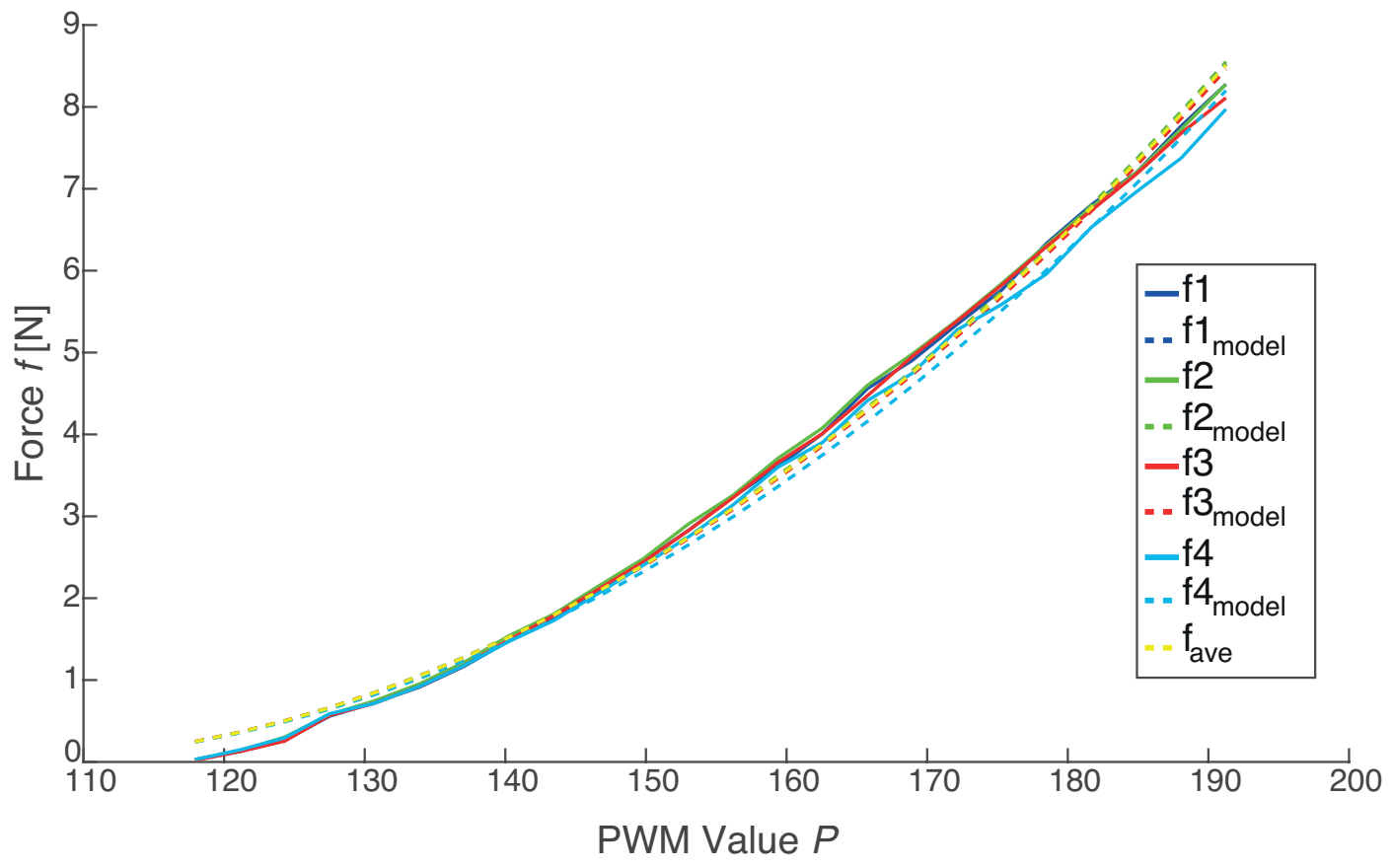

Figure 6: Propeller Thrust as a Function of PWM Command Value

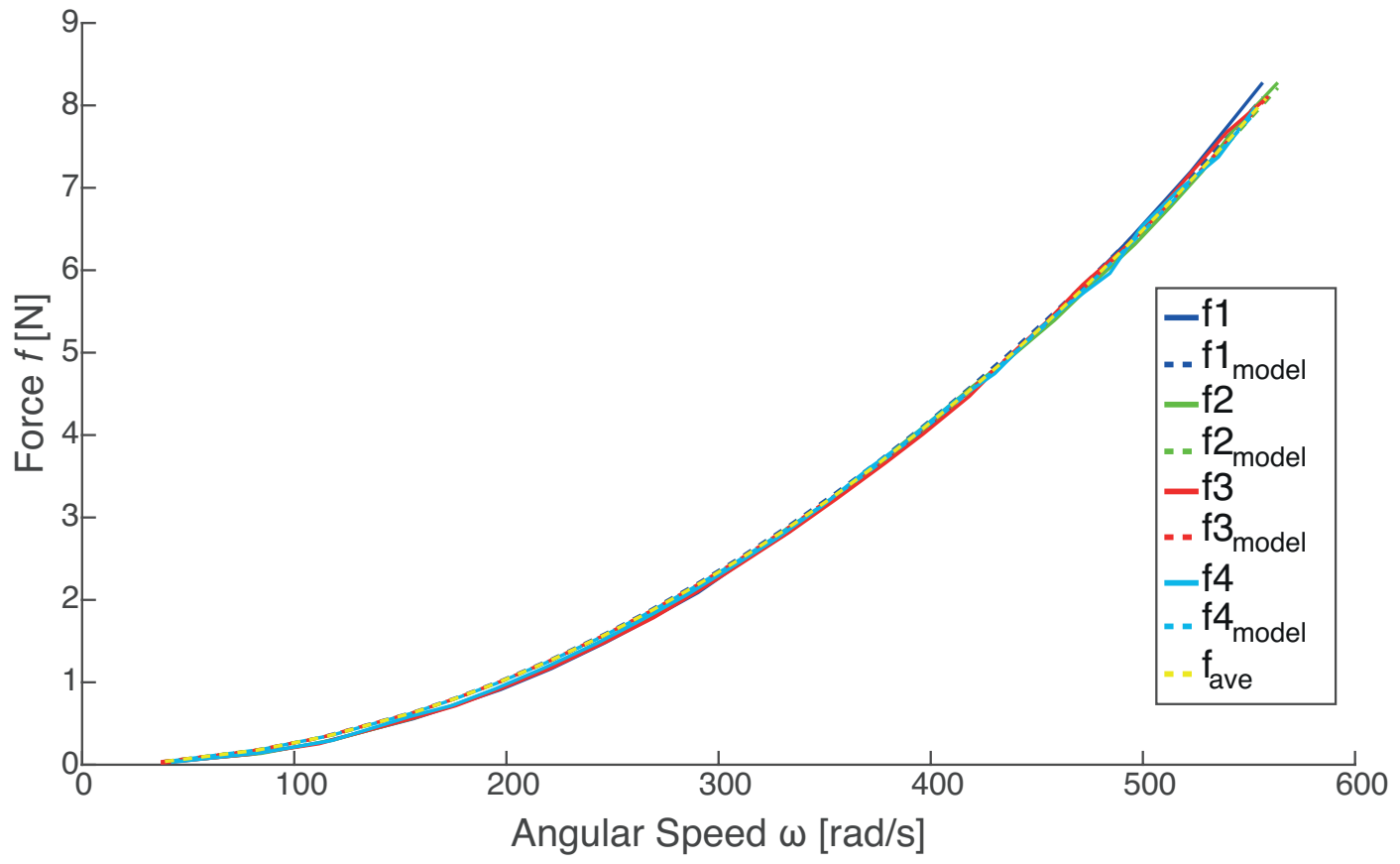

Figure 7: Propeller Thrust as a Function of Propeller Angular Speed

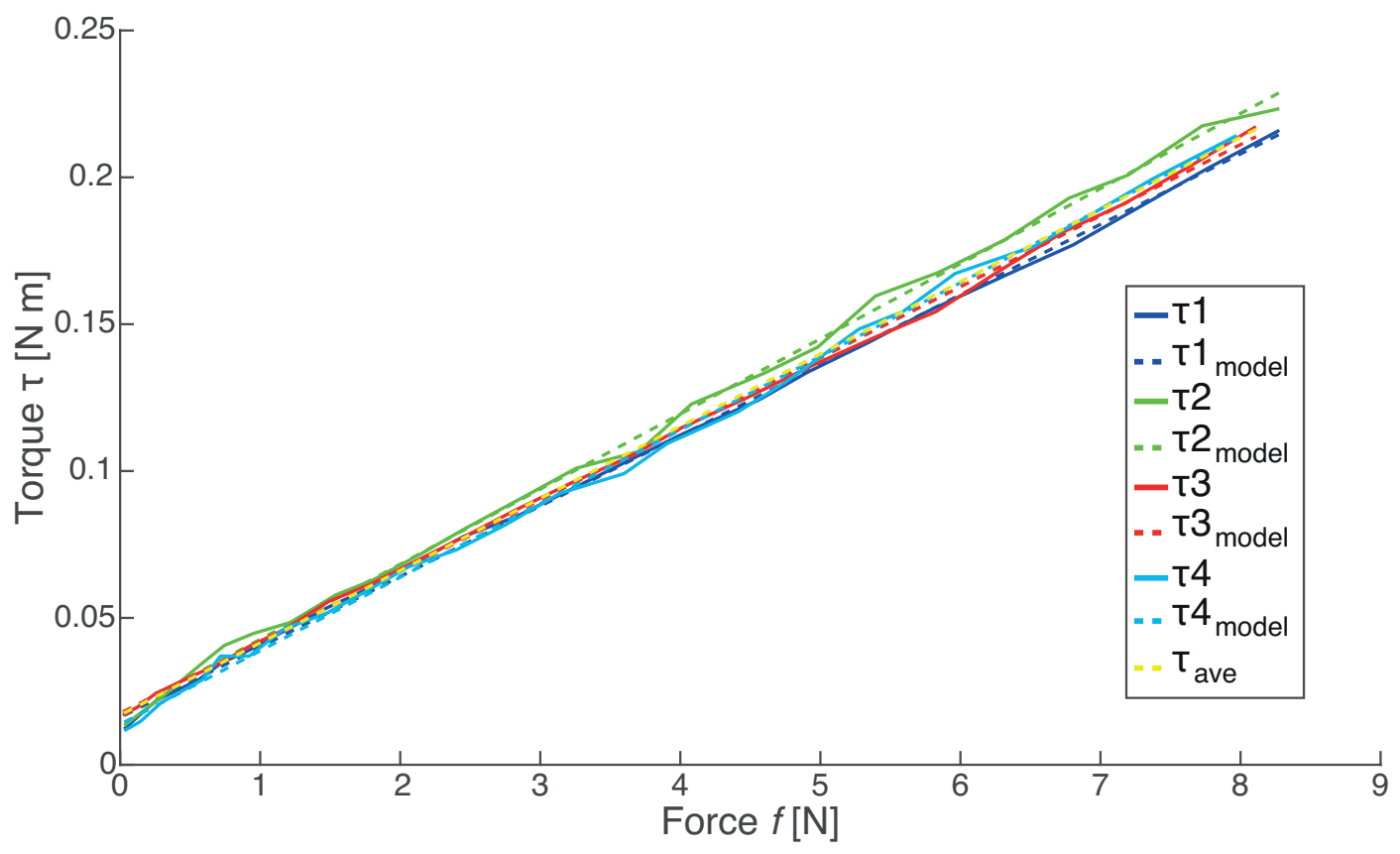

Figure 8: Reaction Torque as a Function of Propeller Thrust

The values of $h_1$, $h_2$, $c_1$, $g_1$, and $g_2$ were estimated from experimental data using the setup in Section 3.2.1. In particular, $h_1$ and $h_2$ were used to convert thrust commands into motor command values. The numbers from 0 to 255 were scaled to 40000 which was the limit of the microcontroller used. It was eventually found out after tuning the fail-safe flight controller in a different work [1] that this improves quadcopter performance. This reduced the discretization of thrust commands by two orders of magnitude and resulted in a highly smoothened or less shaky flight.

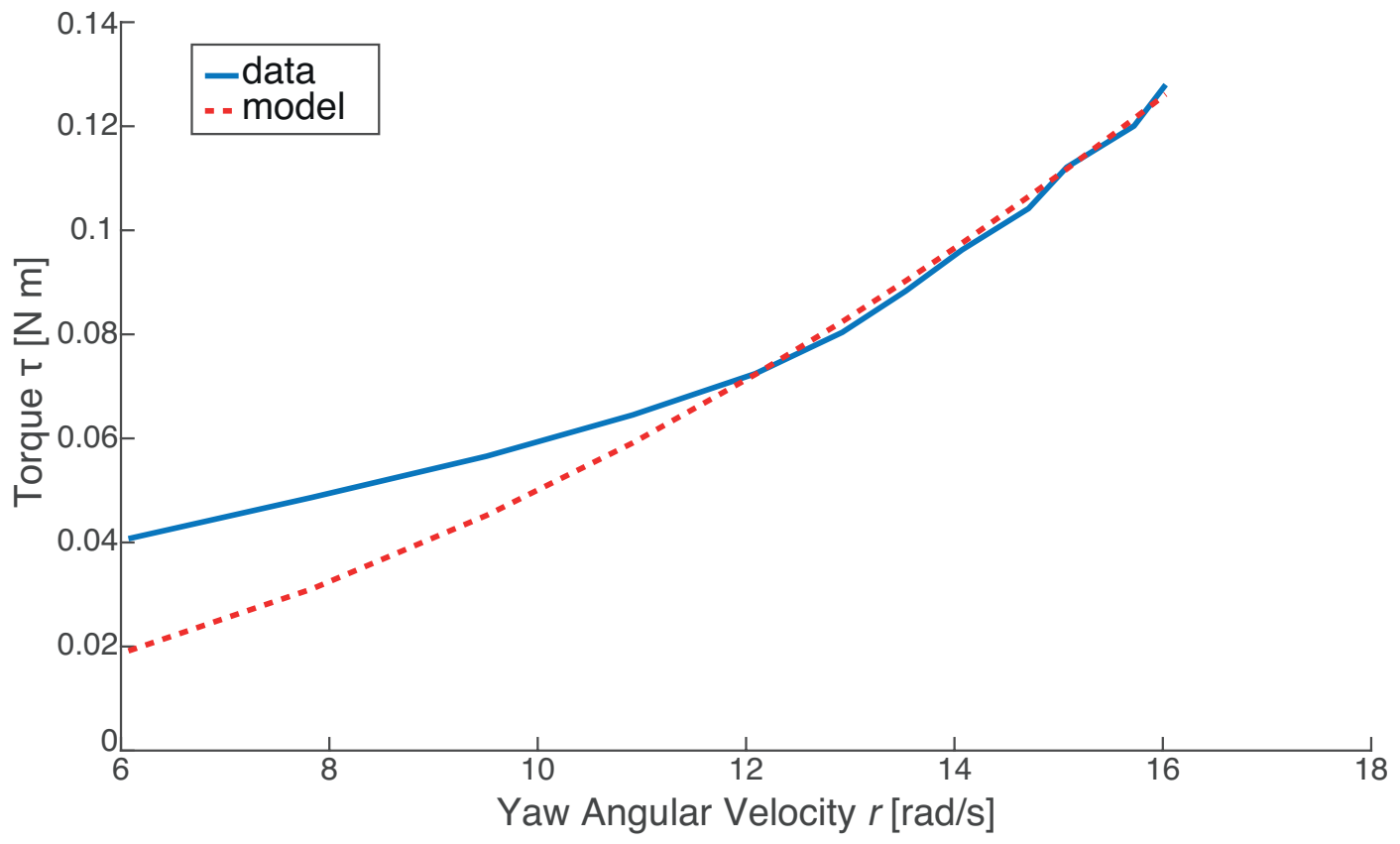

Figure 9: Steady State Yaw Angular Velocities of the Quadcopter Body as a Function of Propeller Torque

Figure 9 shows a plot of steady state angular velocities of the quadcopter body about $\hat{z}^b$ achieved at particular torque values exerted by propeller 1 and propeller 3 as well as a plot of the torque values predicted by the drag coefficients $\gamma_1$ and $\gamma_2$ in Table 2. Notice that the model for the drag is more accurate at higher angular velocities. This property proved useful in fail-safe mode [1] where the quadcopter's yaw rate would exceed normal operation values.

### 4.2. Basic Flight Controller

Figure 10 shows the breakdown of the basic flight controller to various sub-algorithms with their execution times. In the Main Loop, the Timing part updates the time parameters for the PID controller and for the flight instructions. The SPI Angles is the part where angles-related information is requested and received in a buffer. An Adjustable Delay is inserted to control the loop frequency. The Decode Angles part converts the information stored in the buffer into angular information that can be accessed in the code. The Basic Flight Control part is where the PID control law is computed. The Flight Instructions part is where certain flight decisions are made and where the PWM commands are sent to the motors.

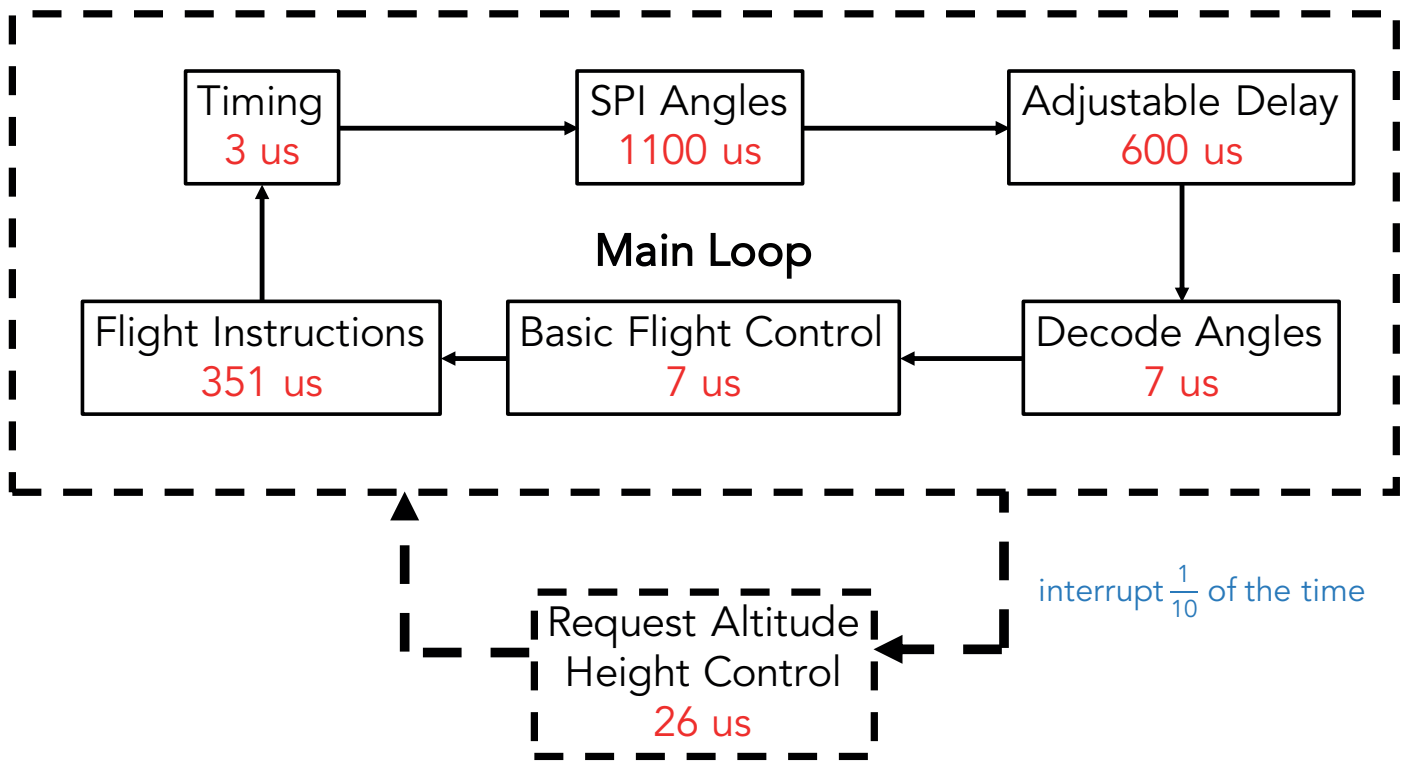

Figure 10: Basic Flight Control

The Main Loop is interrupted every $\frac{1}{10}$ of the time by the Request Attitude part where a function triggers the ultrasonic sensor to begin altitude measurement. Consequently, a Height Control part interrupts every $\frac{1}{10}$ of the time to calculate the requested altitude reading and corresponding altitude PID output. The Main Loop's total execution time was 2068 µs corresponding to a 483.56 Hz frequency while the execution time of the altitude interrupts was 26 µs which occurred at a frequency of around 45 Hz.

The formula for computing the average execution time of the algorithms will now be discussed. Consider algorithm 1 with execution time $T_1$ interrupted by algorithm 2 with execution time $T_2$ every $t_0$ as illustrated in Figure 11.

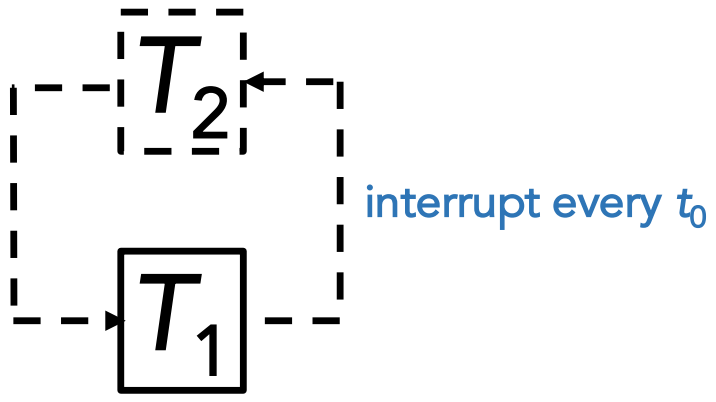

Figure 11: Algorithm 1 Interrupted by Algorithm 2 Every $t_0$

Then the processes can happen as demonstrated in Figure 12.

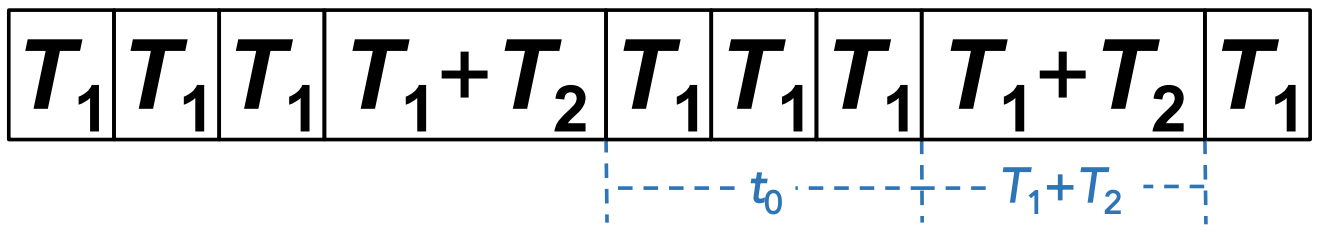

Figure 12: An Example of an Execution of the Processes in Figure 11

The average loop execution time is thus

$$T_{loop}=\frac{t_0}{t_0+T_1+T_2}T_1+\frac{T_1+T_2}{t_0+T_1+T_2}(T_1+T_2) \qquad (29)$$

Using Equation (29) with $t_0=\frac{1}{45\text{ Hz}}$, $T_1$=2068 µs, and $T_2$=26 µs, the average loop period is 2070.23899911 µs so that the average loop frequency of the basic flight controller will be 483 Hz. The observed 450 Hz to 470 Hz main loop frequency in the implementation was due to intermittent drops in the clock frequency of the Raspberry Pi's processor.

### *4.3. Basic Flight Simulation*

This section documents the simulation of basic flight for the modeled quadcopter. As shown in Table 3, the PID outputs were capped based on the characteristic quantity $B$ discussed in Section 3.3.

Table 3: Limits of Output Variables for Basic Flight

| Output Variable | Minimum | Maximum |
|---|---|---|
| $f_1$ | 0 | $2B$ |
| $f_2$ | 0 | $2B$ |
| $f_3$ | 0 | $2B$ |
| $f_4$ | 0 | $2B$ |
| $u_z$ | $-B$ | $B$ |
| $u_\phi$ | $-B$ | $B$ |
| $u_\theta$ | $-B$ | $B$ |
| $u_\psi$ | $-B$ | $B$ |

The basic flight simulations were done through MATLAB R2017a's ode45() solver using the parameters modeled and the experimentally determined drag coefficients in Table 3. In the absence of the bearing system that was later used in the fail-safe implementation [1], the rotational kinetic friction contribution of the bearing $\gamma_2$ was set to zero. These parameters corresponded to the quadcopter implemented.

The state variables of interest are $\phi$, $\theta$, $\psi$, and $z$. The tuned PID gains of the quadcopter in the simulation are listed in Table 4. The second symbol in the subscript indicates the variable to control. Since for this project's basic flight requirement, small steady state errors are of no concern, the integral gain was zeroed out for all variables to avoid unwanted system oscillations. The altitude was first tuned with the remaining variables' PID controllers disabled. Then with only the tuned altitude controller enabled, the roll and pitch controllers were separately tuned. Finally, the yaw controller was tuned with the rest of the tuned controllers enabled.

Table 4: Tuned PID Gains for Basic Flight Simulation

| PID Parameter | Value | Unit |
|---|---|---|
| $k_{p,z}$ | 1.9 | N/m |
| $k_{d,z}$ | 1.6 | (N s)/m |
| $k_{p,\phi}$ | 17.1 | N/rad |
| $k_{d,\phi}$ | 1.3 | (N s)/rad |
| $k_{p,\theta}$ | 17.2 | N/rad |
| $k_{d,\theta}$ | 1.3 | (N s)/rad |
| $k_{p,\psi}$ | 7.7 | N/rad |
| $k_{d,\psi}$ | 2.7 | (N s)/rad |

Results of the simulation with the tuned PID gains are shown in Figure 13, Figure 14, Figure 15, and Figure 16. Figure 13 demonstrates the quadcopter's stability and altitude response to ascent and landing commands. Figure 14 demonstrates the roll, pitch, yaw, and altitude stability of the modeled quadcopter as it accomplishes a positive roll command. The model was also tested for negative roll, and behaved similarly stable. Figure 15 demonstrates the quadcopter's stable response to a negative pitch command. The model was also tested for positive pitch and behaved similarly stable. The quadcopter remains stable and achieves the desired values as expected. Finally, Figure 16 demonstrates both positive and negative yaw responses. As expected, the positive and negative responses in all variables are symmetric, and the attitude and altitude controls are capable of maintaining stability upon a command change in a single flight parameter.

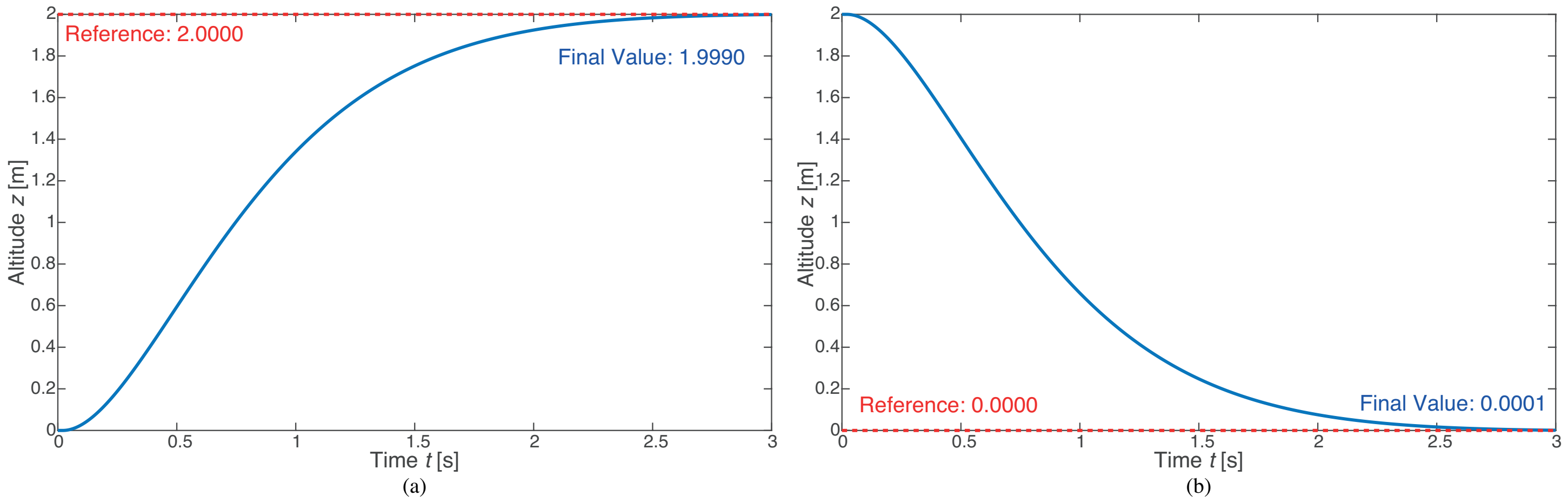

Figure 13: Simulated Quadcopter Altitude Response. In (a), the quadcopter is commanded to ascend from 0 m to 2 m. In (b), the quadcopter is commanded to perform controlled landing from 2 m to 0 m. The quadcopter's attitude was unaffected during ascent or landing.

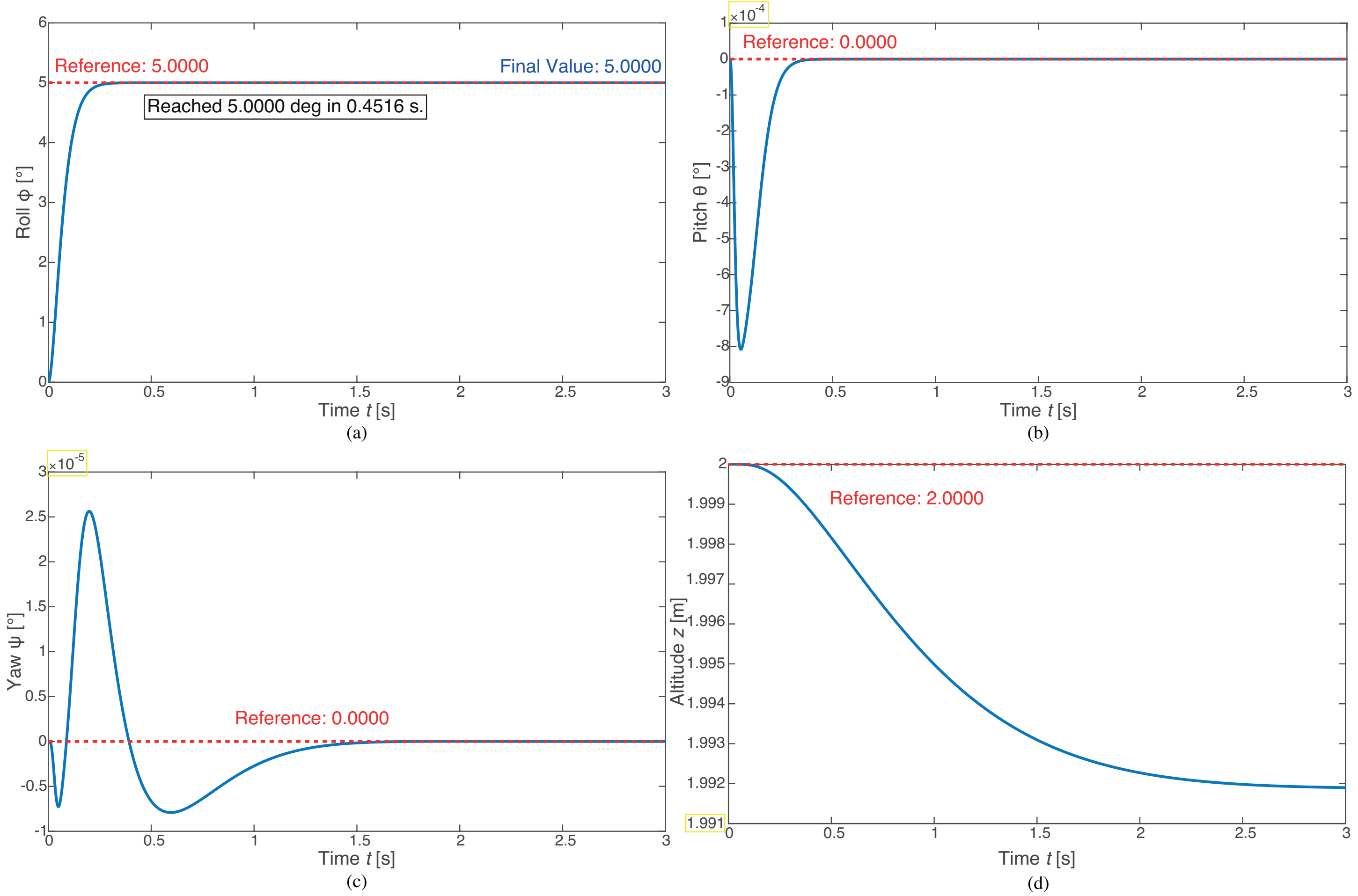

Figure 14: Simulated Quadcopter Response when Banking from 0° to 5° Along the Roll Axis. In (a), the quadcopter's roll angle, initially 0°, smoothly achieves the 5° command. In (b), the variation in the quadcopter's pitch angle with time while banking is negligibly small. In (c), the variation in the quadcopter's yaw angle with time while banking is similarly negligible. In (d), the gradual drop in the quadcopter's altitude while banking is also negligible.

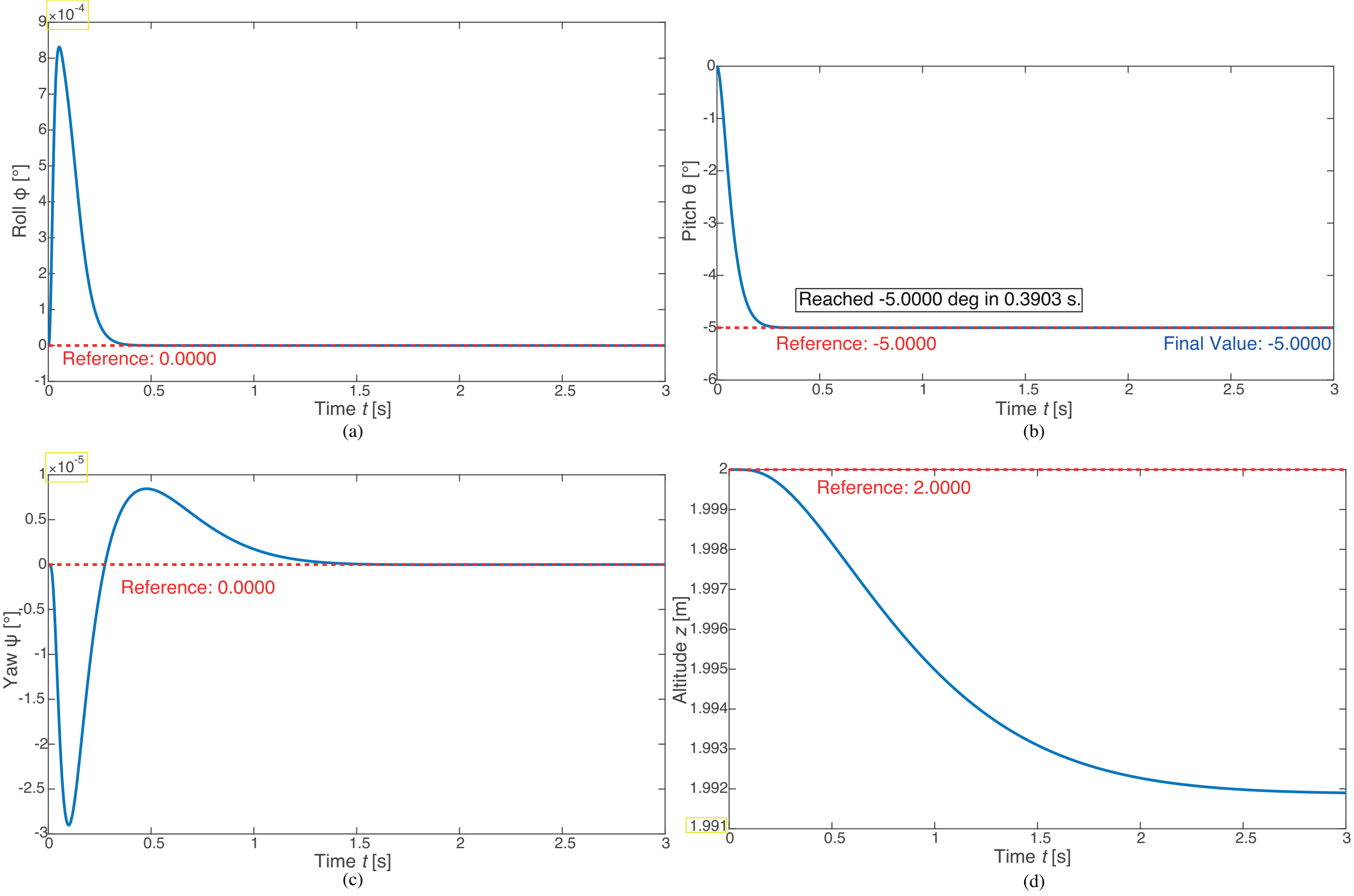

Figure 15: Simulated Quadcopter Response when Banking from 0° to 5° Along the Pitch Axis. In (a), the variation in the quadcopter's roll angle with time while banking is negligibly small. In (b), the quadcopter's pitch angle, initially 0°, smoothly achieves the –5° command. In (c), the variation in the quadcopter's yaw angle with time while banking is similarly negligible. In (d), the gradual drop in the quadcopter's altitude while banking is also negligible.

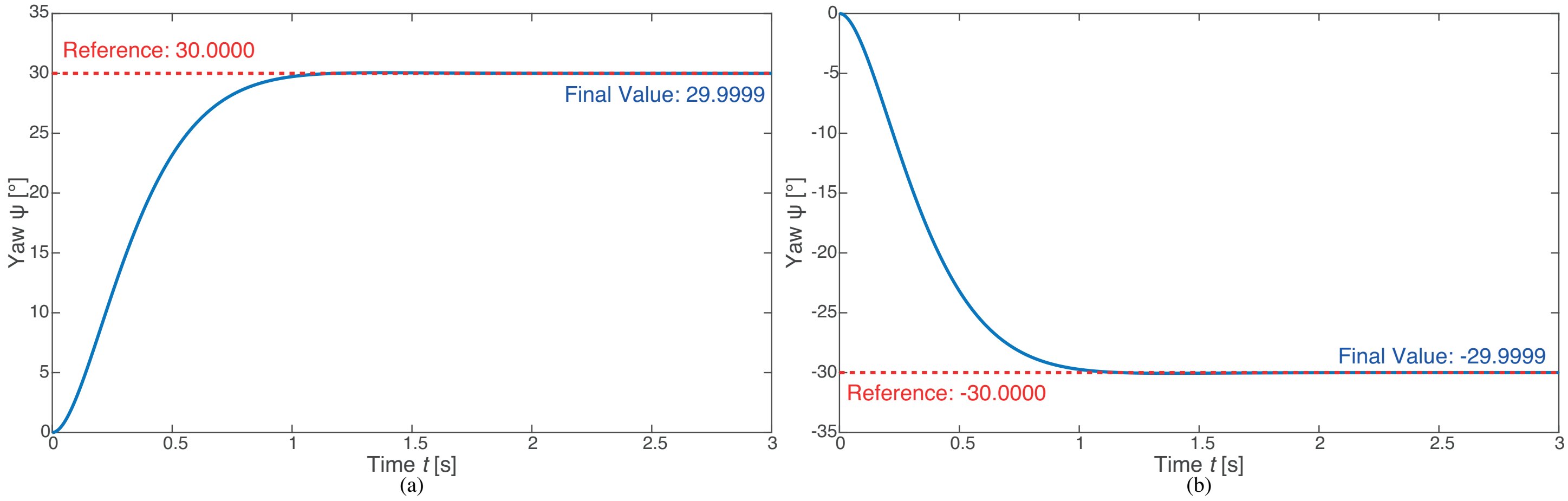

Figure 16: Simulated Quadcopter Response when Rotating ±30° Along the Yaw Axis. In (a), the quadcopter's yaw angle, initially 0°, smoothly achieves the +30° command. In (b), the quadcopter's yaw angle, initially 0°, smoothly achieves the –30° command. The quadcopter's roll, pitch, and altitude were unaffected during rotation.

### 4.4. Basic Flight Implementation

This section documents the simulation of basic flight for the modeled quadcopter.

#### 4.4.1. Control and Actuation Frequency

As simulated, the average loop frequency of the basic flight controller was around 450 Hz with the altitude controller interrupting every $\frac{1}{10}$ of the time. This was found to work well with the ultrasonic proximity sensor having a working frequency of 40 Hz in its datasheet [21]. The frequency used in the motor commands for the basic flight was 449 Hz since the ESC command frequency limit was 450 Hz.

#### 4.4.2. Sensor Fusion Technique

The roll and pitch of the quadcopter can be extracted from the $x$, $y$, and $z$ accelerometer axial readings using [22]

$$\phi=\arctan\left(\frac{accel_y}{accel_z}\right) \tag{30}$$

$$\theta=\arctan\left(\frac{-accel_x}{\sqrt{accel_y{}^2+accel_z{}^2}}\right) \tag{31}$$

while the yaw can be extracted from the $x$, $y$, and $z$ magnetometer axial readings together with the estimates for the roll and the yaw using [23, 24]

$$\psi=\arctan\left(\frac{magnet_z\sin(\phi)-magnet_y\cos(\phi)}{magnet_x\cos(\theta)+magnet_y\sin(\theta)\sin(\phi)+magnet_z\sin(\theta)\cos(\phi)}\right) \tag{32}$$

Although these are absolute information about the attitude, these are very noisy estimates because of the noisy accelerometer and magnetometer readings. The gyroscope readings which are used to compute $\dot{\phi}$, $\dot{\theta}$, and $\dot{\psi}$ are less noisy but they can only give differential estimates for the angles. To get accurate and clean enough attitude estimates, complementary filter may be used to fuse the absolute but noisy estimates with the less noisy but differential estimates according to the weighting equation:

$$\phi=(1-\alpha)\phi_{accel,mag}+\alpha\phi_{gyro} \tag{33}$$

$$\theta=(1-\alpha)\theta_{accel,mag}+\alpha\theta_{gyro} \tag{34}$$

$$\psi=(1-\alpha)\psi_{accel,mag}+\alpha\psi_{gyro} \tag{35}$$

with $\alpha$ as the parameter.

The result is a low-pass filtering of the accelerometer and magnetometer's absolute estimates to remove the noise and a high-pass filtering of the gyroscope's differential estimates to remove the drift which worsens with each iteration. The time constant $\tau=\frac{\alpha dt}{1-\alpha}$ is for both filters for a loop frequency given by $dt$ [25]. The tuning of $\alpha$ was done based on the value of $\tau$ which was adjusted in powers of 2 of the average loop period $dt$ [1].

#### 4.4.3. EMA Filter for Gyroscope

Using the smoothing factor, $\alpha_{EMA}$, an exponential moving average (EMA) filter was applied on the gyroscope readings $p$, $q$, and $r$ in order to remove noise according to the formula

$$p_i=\alpha_{EMA,p}p_i+\left(1-\alpha_{EMA,p}\right)p_{i-1} \tag{36}$$

$$q_i=\alpha_{EMA,q}q_i+\left(1-\alpha_{EMA,q}\right)q_{i-1} \tag{37}$$

$$r_i=\alpha_{EMA,r}r_i+\left(1-\alpha_{EMA,r}\right)r_{i-1} \tag{38}$$

The time constant of an exponential moving average is the amount of time for the smoothened unit step response to reach $1-{}^1/_e\approx 63.2\%$. The relationship between this time constant, $\tau_{EMA}$, and $\alpha_{EMA}$ is given by

$$\alpha_{EMA}=1-e^{\frac{-dt}{\tau_{EMA}}} \tag{39}$$

where $dt$ is the iteration frequency [26].

#### 4.4.4. Sensor Tuning

The exponential smoothing of $p$, $q$, $r$ were done first before performing the sensor fusion technique [1]. The $\alpha_{EMA}$ for each gyroscope reading was tuned while the quadcopter was being held mid-air. The tuning of $\alpha_{EMA}$ was done based on the value of $\tau_{EMA}$, in powers of 2 of the average loop period $dt$ [1]. $\tau_{EMA}$ for each of the variables was doubled until the amplitude of oscillations in $p$ and $q$ were within 0.01 $\frac{\text{rad}}{\text{s}}$. The same $\alpha_{EMA}$ was adopted for $r$. Since the $r$ readings of the sensor was inherently less noisy, the amplitude oscillations in $r$ using the same $\alpha_{EMA}$ in $p$ and $q$ were within 0.006 $\frac{\text{rad}}{\text{s}}$.

After tuning the value of $\alpha_{EMA}$, the quadcopter was made to sit still on a flat surface for 1 minute. The gyroscope readings along each rotation axis for the minute were averaged to get the gyroscopic axial bias which must be subtracted from the future $p$, $q$, $r$ readings of the gyroscope. The corresponding angular values from the accelerometer readings for the minute were also averaged to get the biases on the estimates of $\phi$ and $\theta$; these must be subtracted from the future accelerometer estimates of $\phi$ and $\theta$ prior to sensor fusion. Meanwhile, the $\alpha$ value for the complementary filters in the roll, pitch, and yaw were increased until the amplitude of the oscillations were within ±0.1°. Table 5 lists the filter values.

Table 5: EMA and Complementary Filter Values

| Gyroscopic Axis | $\alpha_{EMA}$ |
|---|---|
| $p$ | 0.007782062 |
| $q$ | 0.007782062 |
| $r$ | 0.007782062 |
| Angular Variable | $\alpha$ |
| $\phi$ | 0.992248062 |
| $\theta$ | 0.992248062 |
| $\psi$ | 0.984615385 |

#### *4.4.5. PID Tuning*

A lot of non-idealities that appeared in the actual implementation of basic flight were not captured in the simulation. These include irregular and intermittent delays in the processor, ESCs, and actuation of motors in addition to sensor noise and possibly slight modeling discrepancies. Thus, although implementing the simulated controller gains worked for the quadcopter's basic flight, its performance was improved by altering the simulated tuned controller gain values to better adapt to these non-idealities. Using the simulated gains in Table 4 as the starting point, the basic flight controller PID gains were retuned to the values listed in Table 6. Note that small integral gains for the roll and pitch angles, which were zeroed out in the simulation, became necessary in the implementation to ensure that the quadcopter did not drift from the reference command values over time, since drifting along the roll and pitch axes would quickly accelerate the quadcopter to dangerously high translational speeds.

Table 6: Tuned PID Gains for Basic Flight Implementation

| PID Parameter | Value | Unit |
|---|---|---|
| $k_{p,z}$ | 2.4 | N/m |
| $k_{d,z}$ | 0.14 | (N s)/m |
| $k_{p,\phi}$ | 4.48 | N/rad |
| $k_{d,\phi}$ | 0.221 | (N s)/rad |
| $k_{i,\phi}$ | 0.0045 | N/(rad s) |
| $k_{p,\theta}$ | 4.12 | N/rad |
| $k_{d,\theta}$ | 0.13 | (N s)/rad |
| $k_{i,\theta}$ | 0.0045 | N/(rad s) |
| $k_{p,\psi}$ | 7.55 | N/rad |
| $k_{d,\psi}$ | 0.248 | (N s)/rad |

To avoid external airflow disturbances such as wind gusts during the implementation of basic flight, the quadcopter was flown indoors. The hovering test produced the altitude, roll, pitch, and yaw response in Figure 17. In the following plots, the actual response is shown in blue whereas the reference command is shown in red. As expected, because of the aforementioned non-idealities, more jitter is present in the actual quadcopter response than in the ideal simulation results, but the settling of each parameter emphasizes the correct controller design which in turn verifies the quadcopter model derived.

Compared with the obtained altitude response using the flight controller in this work, the altitude control performance of typical modern UAVs [27] are more than twice as jittery. Moreover, the altitude tracking error of modern quadcopters using a similar ultrasonic sensor [28] can be more than 20 times worse than that in Figure 17. Although quadcopters using more sophisticated sensors like laser radar [29] can reduce altitude tracking error to almost one order of magnitude better than the altitude response presented in this paper, the controller in this work nevertheless performs more smoothly and with an almost equal fluctuation in the altitude.

Moreover, the graph responses for positive and negative banking along the roll and pitch directions are shown in Figure 18 and Figure 19. The quadcopter was set to hover at the beginning of both tests, and upon achieving stability, commanded to bank before returning to hover mode. In both plots, the quadcopter was able to find and settle around the command values in each change of state. The yaw responses to positive and negative rotation commands as shown in Figure 20 also demonstrate quick stabilization in either direction.

In terms of attitude control, the roll, pitch, and yaw responses of the quadcopter system developed in this work are typically more jittery compared with the responses of quadcopters using popular firmware systems like ArduPilot [30] and Pixhawk [31, 32] but the developed system can nonetheless achieve a lower average tracking error in the roll and pitch angles.

## 5. Summary and Conclusion

The study came up with alternative methods to determine thrust vs. PWM command, thrust vs. angular speed, and reaction torque vs. thrust relationships of the quadcopter's propellers as well as the coefficients of the aerodynamic drag acting on the quadcopter, thereby completing a dynamic model for the quadcopter. With the specific model, the study extended into designing and simulating a basic flight controller. The simulation led to determining PID values and limits which served as starting points for the actual controller implemented. Plots of the simulated responses to thorough actual tests demonstrated the expected stable flight of the quadcopter. These results further verified the completeness and correctness of the model obtained. This method was repeated for multiple versions of quadcopter with a completely different set of motors, ESCs, and propellers. Using the same methodology, corresponding values were obtained and similar behavior was demonstrated.

The quadcopter implemented in this paper was used in another work [1] which is the implementation of a fail-safe controller for a quadcopter that experiences a motor failure. A working basic flight controller is a stepping stone to solving the motor failure problem.

## Conflict of Interest

The authors declare no conflict of interest.

## Acknowledgment

This project was supported in part by grants from the UP Engineering Research & Development Foundation, Inc. Publication of this paper was made possible by the Engineering Research and Development for Technology program of the Department of Science and Technology in the Philippines.

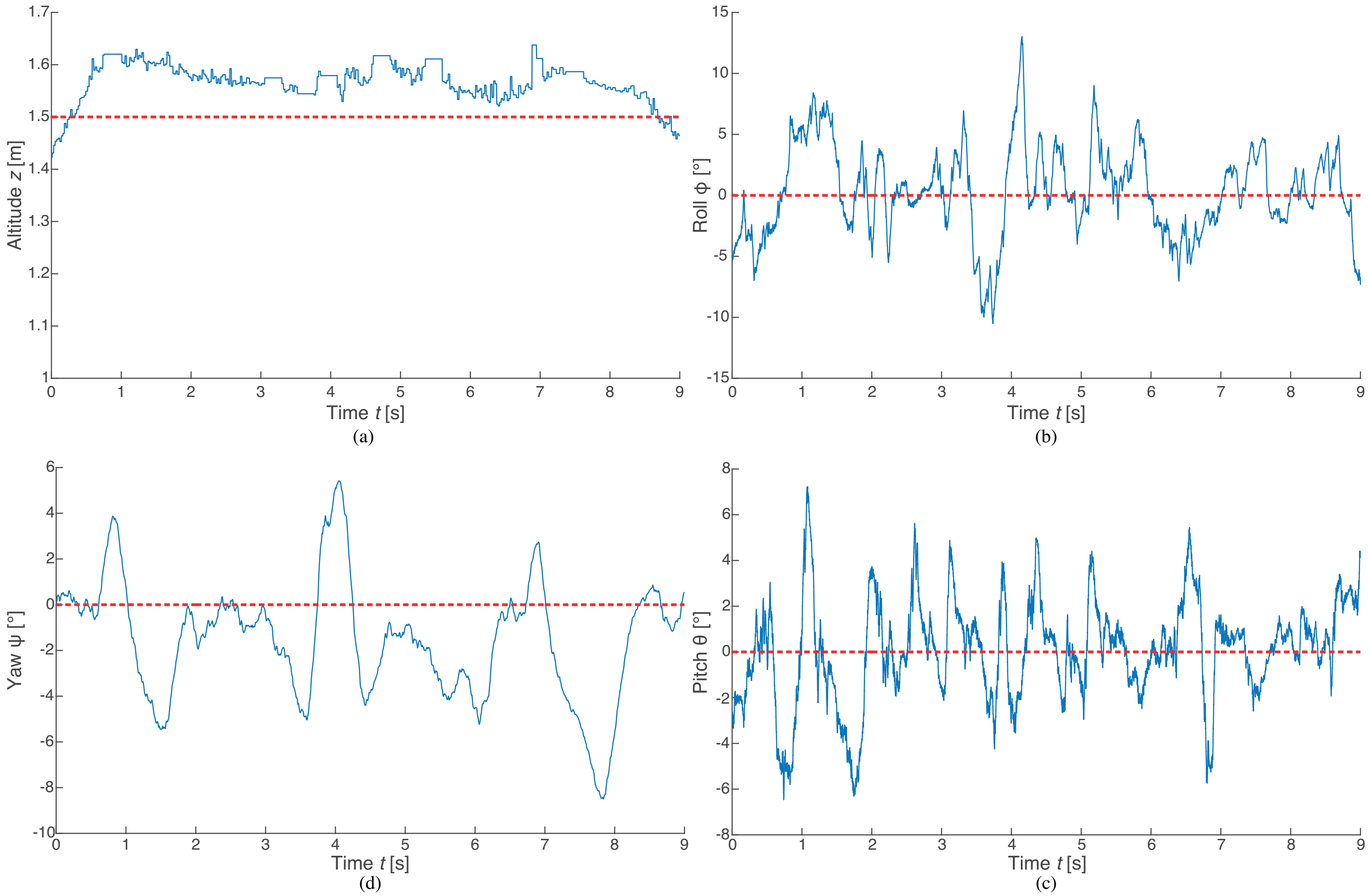

Figure 17: Quadcopter Hover Response. In (a), the quadcopter's altitude remains stably close to the 1.5 m reference value with time. In (b), the quadcopter's roll angle remains stably close to the 0° reference value with time. In (c), the quadcopter's pitch angle also remains stably close to the 0° reference value with time. In (d), the quadcopter's yaw angle remains similarly stable and close to the 0° reference value with time.

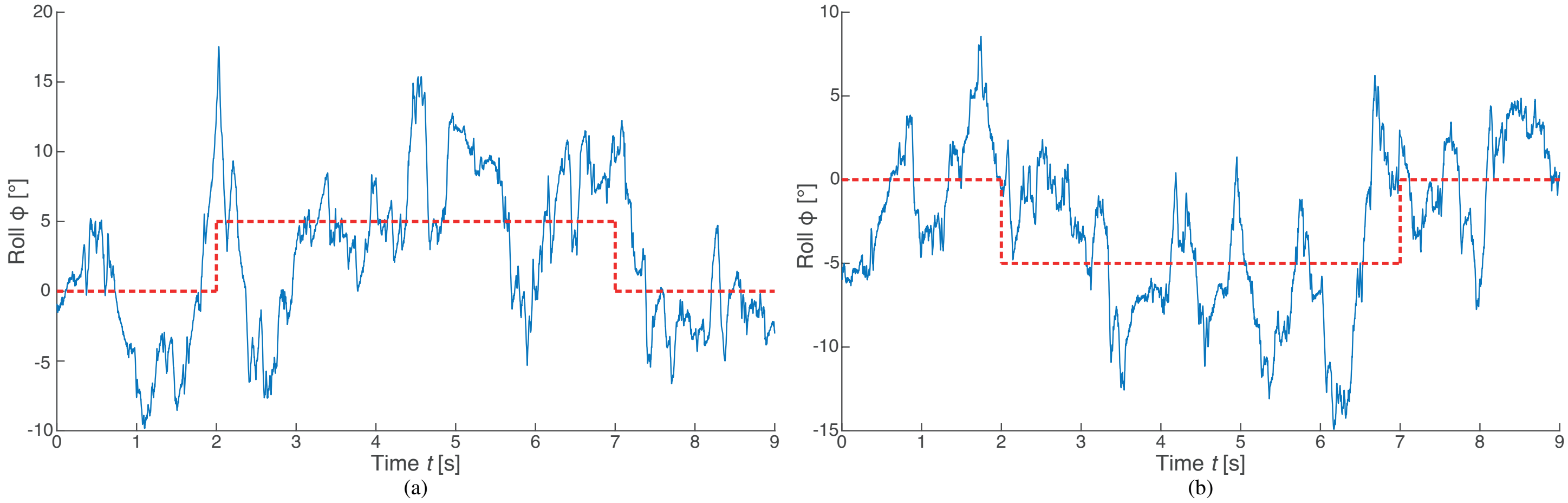

Figure 18: Quadcopter Roll Response to Positive and Negative Banking Commands. In (a), the quadcopter's roll angle, initially 0°, achieves a +5° command for 3 s before returning to 0°. In (b), the quadcopter's roll angle, initially 0°, achieves a –5° command for 5 s before returning to 0°. The quadcopter's pitch, yaw, and altitude were not significantly affected during banking.

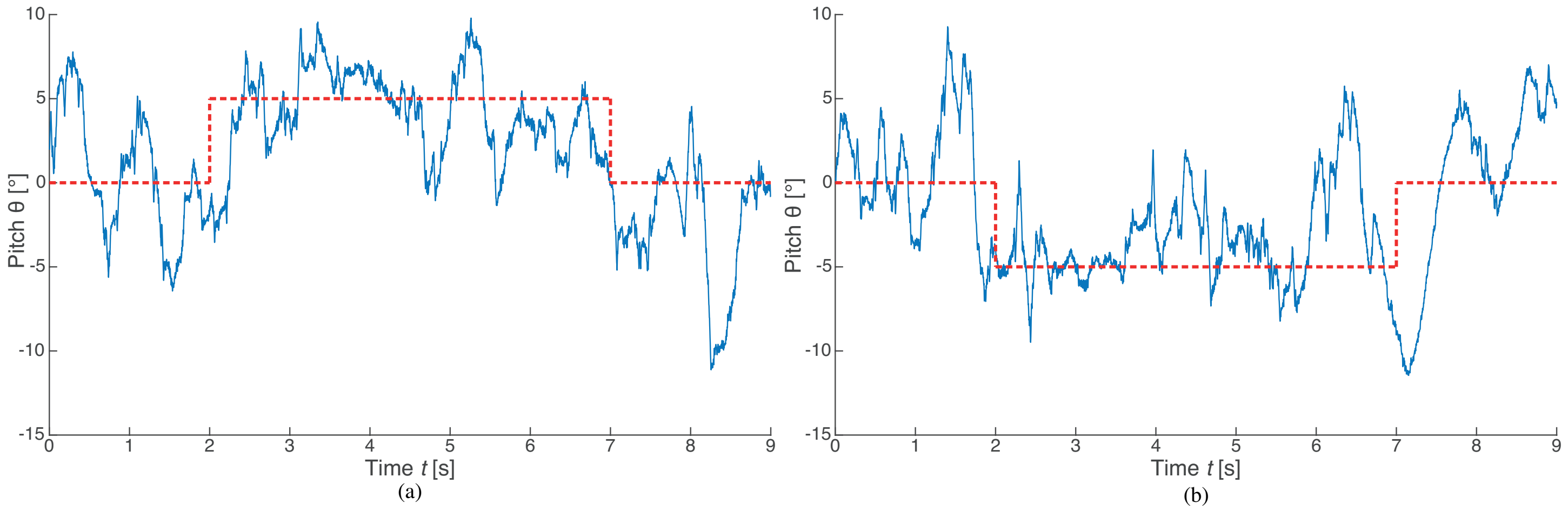

Figure 19: Quadcopter Pitch Response to Positive and Negative Banking Commands. In (a), the quadcopter's pitch angle, initially 0°, achieves a +5° command for 3 s before returning to 0°. In (b), the quadcopter's pitch angle, initially 0°, achieves a –5° command for 5 s before returning to 0°. The quadcopter's roll, yaw, and altitude were not significantly affected during banking.

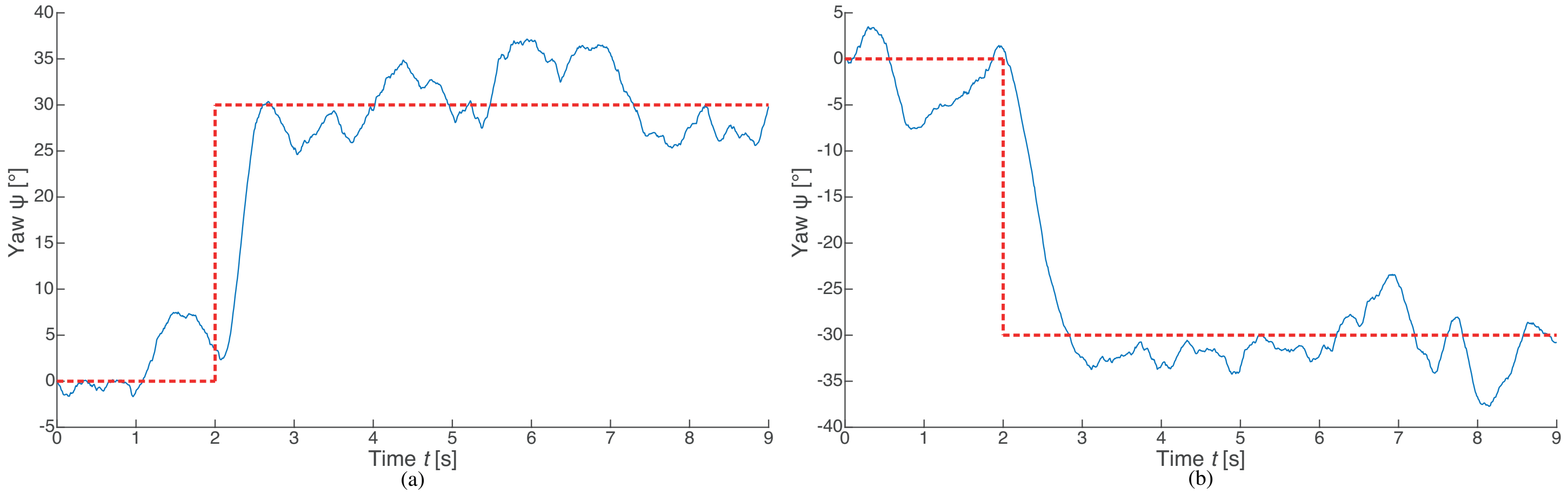

Figure 20: Quadcopter Yaw Response to Positive and Negative Rotation Commands. In (a), the quadcopter's yaw angle, initially 0°, achieves the +30° command. In (b), the quadcopter's yaw angle, initially 0°, achieves the –30° command. The quadcopter's roll, pitch, and altitude were not significantly affected during rotation.

## References

[1] G. P. S. Rible, N. A. A. Arriola, M. C. Ramos, "Fail-Safe Controller Architectures for Quadcopter with Motor Failures" in 2020 6th International Conference on Control, Automation and Robotics (ICCAR), Singapore, 2020. https://doi.org/10.1109/ICCAR49639.2020.9108038

[2] B. Fan, J. Sun, Y. Yu, "A LQR controller for a quadrotor: Design and experiment" in 2016 31st Youth Academic Annual Conference of Chinese Association of Automation (YAC), Wuhan, Hubei, China, 2016. https://doi.org/10.1109/YAC.2016.7804869

[3] J. Qiao, Z. Liu, Y. Zhang, "Gain scheduling PID control of the quad-rotor helicopter" in 2017 IEEE International Conference on Unmanned Systems (ICUS), Beijing, China, 2017. https://doi.org/10.1109/ICUS.2017.8278414

[4] Y. Sen, W. Zhongsheng, "Quad-Rotor UAV Control Method Based on PID Control Law" in 2017 International Conference on Computer Network, Electronic and Automation (ICCNEA), Xi'an, Shaanxi, China, 2017. https://doi.org/10.1109/ICCNEA.2017.24

[5] R. W. Beard, "Quadrotor Dynamics and Control," Tech. Rep., Brigham Young University, 2008.

[6] M. W. Mueller, R. D'Andrea, "Stability and control of a quadrocopter despite the complete loss of one, two, or three propellers" in 2014 IEEE International Conference on Robotics and Automation (ICRA), Hong Kong, China, 2014. https://doi.org/10.1109/ICRA.2014.6906588

[7] G. K. Batchelor, An Introduction to Fluid Dynamics, Cambridge University Press, 2000.

[8] Y. Nakayama, R. Boucher, "9 - Drag and lift" in Introduction to Fluid Mechanics, Butterworth-Heinemann, 1998.

[9] A. Nagaty, S. Saeedi, C. Thibault, M. Seto, H. Li, "Control and Navigation Framework for Quadrotor Helicopters" J. Intell. Robot. Syst., **70**(1-4), 1–12, 2013. https://doi.org/10.1007/s10846-012-9789-z

[10] H. t. M. N. ElkHoly, "Dynamic Modeling and Control of a Quadrotor Using Linear and Nonlinear Approaches," MS Thesis, The American University in Cairo, 2014.

[11] DJI E600 Multirotor Propulsion System, User Manual V1.04, DJI Technology Co., Ltd., 2014.

[12] M.-G. Yoon, "On Driving Signal of Electronic Speed Controller for Small Multi-Rotor Helicopter" Int. J. Eng. Res. Technol., **4**(11), 2278–0181, 2015. https://www.ijert.org/on-driving-signal-of-electronic-speed-controller-for-small-multi-rotor-helicopter

[13] D. M. Filatov, A. V. Devyatkin, A. I. Friedrich, "Quadrotor parameters identification and control system design" in 2017 IEEE Conference of Russian Young Researchers in Electrical and Electronic Engineering (EIConRus), St. Petersburg, Russia, 2017. https://doi.org/10.1109/EIConRus.2017.7910684

[14] D. M. Filatov, A. I. Friedrich, A. V. Devyatkin, "Parameters identification of thrust generation subsystem for small unmanned aerial vehicles" in 2017 IEEE II International Conference on Control in Technical Systems (CTS), St. Petersburg, Russia, 2017. https://doi.org/10.1109/CTSYS.2017.8109574

[15] T. Bresciani, "Modelling, Identification and Control of a Quadrotor Helicopter," MS Thesis, Lund University, 2008.

[16] A. Hughes, Electric Motors and Drives: Fundamentals, Types and Applications, 3rd ed., Elsevier Science/Newnes, 2005.

[17] H. Young, R. Freedman, University Physics with Modern Physics, 14th (Global) ed., Pearson Education, 2016.

[18] Y. Nakayama, R. Boucher, Introduction to Fluid Mechanics, Butterworth-Heinemann, 1998.

[19] J. Shigley, J. Uicker, Theory of Machines and Mechanisms, International ed., McGraw-Hill, 1980.

[20] A. H. Ahmed, A. N. Ouda, A. M. Kamel, Y. Z. Elhalwagy, "Attitude stabilization and altitude control of quadrotor" in 2016 12th International Computer Engineering Conference (ICENCO), Cairo, Egypt, 2016. https://doi.org/10.1109/ICENCO.2016.7856456

[21] Ultrasonic Ranging Module HC - SR04, User's Manual, ElecFreaks, n. d.

[22] M. Pedley, "Tilt Sensing Using a Three-Axis Accelerometer," Freescale Semiconductor Application Note, **1**, 2012–2013, NXP Semiconductors N.V., 2013.

[23] H. Vathsangam, "Magnetometer - My IMU estimation experience," 2010. https://sites.google.com/site/myimuestimationexperience/sensors/magnetometer

[24] "3-Axis Compass Sensor Set HMC1055" in Sensing Earth's magnetic field: Compassing, Magnetometry and Dead Reckoning Solutions, Magnetic Sensors Product Catalog, Honeywell Solid State Electronics Center, 2001. https://stevenengineering.com/Tech_Support/PDFs/31MAGSEN.pdf

[25] S. Colton, "The Balance Filter," White Paper, 2007. https://d1.amobbs.com/bbs_upload782111/files_44/ourdev_665531S2JZG6.pdf

[26] G. Stanley, "Exponential Filter," 2010-2020. https://gregstanleyandassociates.com/whitepapers/FaultDiagnosis/Filtering/Exponential-Filter/exponential-filter.htm

[27] W. Liu, C. Yu, X. Wang, Y. Zhang, Y. Yu, "The Altitude Hold Algorithm of UAV Based on Millimeter Wave Radar Sensors" in 2017 9th International Conference on Intelligent Human-Machine Systems and Cybernetics (IHMSC), Hangzhou, Zhejiang, China, 2017. https://doi.org/10.1109/IHMSC.2017.106

[28] D. J. Esteves, A. Moutinho, J. R. Azinheira, "Stabilization and Altitude Control of an Indoor Low-Cost Quadrotor: Design and Experimental Results" in 2015 IEEE International Conference on Autonomous Robot Systems and Competitions, Vila Real, Vila Real District, Portugal, 2015. https://doi.org/10.1109/ICARSC.2015.30

[29] J. Zhao, Y. Li, D. Hu, Z. Pei, "Design on altitude control system of quad rotor based on laser radar" in 2016 IEEE International Conference on Aircraft Utility Systems (AUS), Beijing, China, 2016. https://doi.org/10.1109/AUS.2016.7748029

[30] Ö. Elbir, A. U. Batmaz, C. Kasnakoğlu, "Improving quadrotor 3-axes stabilization results using empirical results and system identification" in 2013 9th Asian Control Conference (ASCC), Istanbul, Turkey, 2013. https://doi.org/10.1109/ASCC.2013.6606281

[31] E. A. Niit, W. J. Smit, "Integration of model reference adaptive control (MRAC) with PX4 firmware for quadcopters" in 2017 24th International Conference on Mechatronics and Machine Vision in Practice (M2VIP), Auckland, New Zealand, 2017. https://doi.org/10.1109/M2VIP.2017.8211479

[32] S. Sakulthong, S. Tantrairatn, W. Saengphet, "Frequency Response System Identification and Flight Controller Tuning for Quadcopter UAV" in 2018 Third International Conference on Engineering Science and Innovative Technology (ESIT), North Bangkok, Thailand, 2018. https://doi.org/10.1109/ESIT.2018.8665114